%% file: main.tex
\documentclass[webpdf,contemporary,large,namedate]{oup-authoring-template}

\usepackage[numbers,sort&compress]{natbib}

\setcitestyle{maxnames=2}
\theoremstyle{thmstyleone}%
\theoremstyle{thmstyletwo}%
\theoremstyle{thmstylethree}%
\newtheorem{definition}{Definition}
\newcommand\modelab{S$2$-SPM }

\usepackage{caption}

\usepackage{subcaption}

\usepackage{bm}
\usepackage{booktabs}
\usepackage{amsmath}
\usepackage{amsfonts}
\usepackage{graphicx}
\usepackage{pdfpages}

\newcommand{\ie}{i.\,e., }

\begin{document}

\journaltitle{Preprint}
\DOI{TBD}
\copyrightyear{2025}
\pubyear{ .}
\access{ .}
\appnotes{.}

\firstpage{1}


\title[The Signed Two-Space Proximity Model]{The Signed Two-Space Proximity Model for Learning Representations in Protein-Protein Interaction Networks}

\author[1,$\ast$]{Nikolaos Nakis\ORCID{0000-0001-9311-3458}}
\author[2]{Chrysoula Kosma\ORCID{0000-0003-4255-4485}}
\author[3]{Anastasia Brativnyk\ORCID{0009-0002-6245-4414}}
\author[1]{Michail Chatzianastasis\ORCID{0000-0002-9905-1646}}
\author[1]{Iakovos Evdaimon\ORCID{0009-0005-2731-571X}}
\author[1]{Michalis Vazirgiannis\ORCID{0000-0001-5923-4440}}

\authormark{Nakis, Kosma, Brativnyk, Chatzianastasis, Evdaimon, Vazirgiannis}
\address[1]{\orgdiv{LIX, \'Ecole Polytechnique}, \orgname{Institute Polytechnique de Paris}, \orgaddress{\street{Rue Honoré d'Estienne d'Orves}, \postcode{91120}, \state{Palaiseau}, \country{France}}}
\address[2]{\orgdiv{Centre Borelli, \'Ecole Normale Superieure (ENS) Paris-Saclay}, \orgname{Université Paris-Saclay}, \orgaddress{\street{Avenue des Sciences}, \postcode{91190}, \state{Gif-sur-Yvette}, \country{France}}}
\address[3]{\orgdiv{Ancient Genomics Laboratory}, \orgname{The Francis Crick Institute}, \orgaddress{\street{Midland Road}, \postcode{NW1 1AT}, \state{London}, \country{United Kingdom}}}

\corresp[$\ast$]{Corresponding authors. \href{email:nicolaos.nakis@gmail.com}{nicolaos.nakis@gmail.com}, \href{email:sissykosm2@gmail.com}{sissykosm2@gmail.com}}




\abstract{\textbf{Motivation:}
Accurately predicting complex protein-protein interactions (PPIs) is crucial for decoding biological processes, from cellular functioning to disease mechanisms. However, experimental methods for determining PPIs are computationally expensive. Thus, attention has been recently drawn to machine learning approaches. Furthermore, insufficient effort has been made toward analyzing signed PPI networks, which capture both activating (positive) and inhibitory (negative) interactions. 
To accurately represent biological relationships, we present the Signed Two-Space Proximity Model (S2-SPM) for signed PPI networks, which explicitly incorporates both types of interactions, reflecting the complex regulatory mechanisms within biological systems. This is achieved by leveraging two independent latent spaces to differentiate between positive and negative interactions while representing protein similarity through proximity in these spaces. Our approach also enables the identification of archetypes representing extreme protein profiles.\\ \textbf{Results:} S2-SPM's superior performance in predicting the presence and sign of interactions in SPPI networks is demonstrated in link prediction tasks against relevant baseline methods. Additionally, the biological prevalence of the identified archetypes is confirmed by an enrichment analysis of Gene Ontology (GO) terms, which reveals that distinct biological tasks are associated with archetypal groups formed by both interactions. This study is also validated regarding statistical significance and sensitivity analysis, providing insights into the functional roles of different interaction types. Finally, the robustness and consistency of the extracted archetype structures are confirmed using the Bayesian Normalized Mutual Information (BNMI) metric, proving the model's reliability in capturing meaningful SPPI patterns. \\
\textbf{Availability:} S2-SPM is implemented and freely available under MIT license at \url{https://github.com/Nicknakis/S2SPM}.}
\maketitle

\input{intro_new}
\input{materials_methods}

\input{results}

\input{discussion}
\vspace{-7pt}
\section*{Data availability}
The original data for creating the signed protein-protein interaction networks are publicly available at the SIGnaling Network Open Resource 3.0 portal: \url{https://signor.uniroma2.it/downloads.php}. The Gene Ontology (GO) terms for the corresponding proteins are publicly available and can be accessed at the \textsc{UniProt} database: \url{https://www.uniprot.org}.
\vspace{-9pt}
\bibliographystyle{abbrvnat}


\bibliography{main.bib}

\includepdf[pages={1-5}]{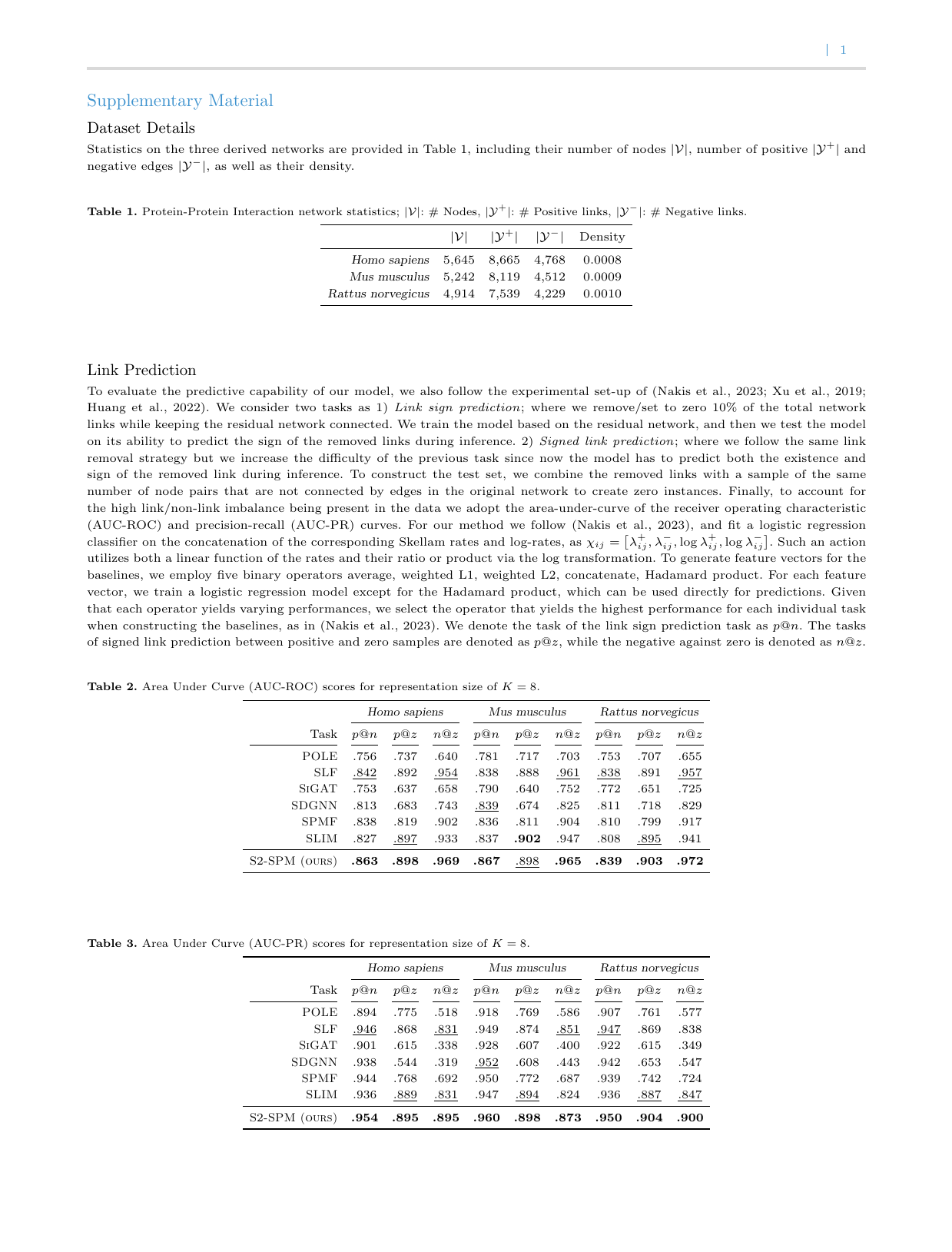}

\end{document}


\section*{Supplementary Material}
\subsection*{Dataset Details}
Statistics on the three derived networks are provided in Table \ref{tab:networks}, including their number of nodes $|\mathcal{V}|$, number of positive $|\mathcal{Y}^{+}|$ and negative edges $|\mathcal{Y}^{-}|$, as well as their density.

\begin{table}[htbp]
\centering
\caption{Protein-Protein Interaction network statistics; $|\mathcal{V}|$: \# Nodes, $ |\mathcal{Y}^+|$: \# Positive links, $|\mathcal{Y}^-|$: \# Negative links.}
\label{tab:networks}
\resizebox{0.4\columnwidth}{!}{%
\begin{tabular}{rcccc}
\toprule
 & $|\mathcal{V}|$ & $|\mathcal{Y}^{+}|$ & $|\mathcal{Y}^{-}|$ & Density \\\midrule
\textsl{Homo sapiens} & 5,645 & 8,665 & 4,768 & 0.0008 \\
\textsl{Mus musculus} & 5,242 & 8,119 & 4,512 & 0.0009 \\
\textsl{Rattus norvegicus} & 4,914 & 7,539 & 4,229 & 0.0010 \\\bottomrule
\end{tabular}%
}
\end{table}

\subsection*{Link Prediction}
To evaluate the predictive capability of our model, we also follow the experimental set-up of \citep{SLIM,slf,pole}. We consider two tasks as 1) \textit{Link sign prediction}; where we remove/set to zero $10\%$ of the total network links while keeping the residual network connected. We train the model based on the residual network, and then we test the model on its ability to predict the sign of the removed links during inference. 2) \textit{Signed link prediction}; where we follow the same link removal strategy but we increase the difficulty of the previous task since now the model has to predict both the existence and sign of the removed link during inference. To construct the test set, we combine the removed links with a sample of the same number of node pairs that are not connected by edges in the original network to create zero instances.  Finally, to account for the high link/non-link imbalance being present in the data we adopt the area-under-curve of the receiver operating characteristic (AUC-ROC) and precision-recall (AUC-PR) curves. For our method we follow \citep{SLIM}, and fit a logistic regression classifier on the concatenation of the corresponding Skellam rates and log-rates, as $\chi_{ij}=\big [\lambda_{ij}^{+},\lambda_{ij}^{-},\log \lambda_{ij}^{+},\log \lambda_{ij}^{-}\big ]$. Such an action utilizes both a linear function of the rates and their ratio or product via the log transformation. To generate feature vectors for the baselines, we employ five binary operators {average, weighted L1, weighted L2, concatenate, Hadamard product}. For each feature vector, we train a logistic regression model except for the Hadamard product, which can be used directly for predictions. Given that each operator yields varying performances, we select the operator that yields the highest performance for each individual task when constructing the baselines, as in \citep{SLIM}.
We denote the task of the link sign prediction task as $p@n$. The tasks of signed link prediction between positive and zero samples are denoted as $p@z$, while the negative
against zero is denoted as $n@z$.

\begin{table*}[h]
\centering
\caption{Area Under Curve (AUC-ROC) scores for representation size of $K=8$.}
\label{tab:auc_roc} 
\resizebox{0.6\textwidth}{!}{%
\begin{tabular}{rcccccccccccccccccccc}\toprule
\multicolumn{1}{l}{} & \multicolumn{3}{c}{\textsl{Homo sapiens}} & \multicolumn{3}{c}{\textsl{Mus musculus}} & \multicolumn{3}{c}{\textsl{Rattus norvegicus}}  \\\cmidrule(rl){2-4}\cmidrule(rl){5-7}\cmidrule(rl){8-10}
\multicolumn{1}{r}{Task} & $p@n$ & $p@z$ & $n@z$ & $p@n$ & $p@z$ & $n@z$ & $p@n$ & $p@z$ & $n@z$   \\\cmidrule(rl){1-1}\cmidrule(rl){2-2}\cmidrule(rl){3-3}\cmidrule(rl){4-4}\cmidrule(rl){5-5}\cmidrule(rl){6-6}\cmidrule(rl){7-7}\cmidrule(rl){8-8}\cmidrule(rl){9-9}\cmidrule(rl){10-10}
\textsc{POLE}    &.756 &.737 &.640 &.781 &.717 &.703 &.753 &.707 &.655 \\
\textsc{SLF} &\underline{.842} &.892 &\underline{.954} &.838 &.888 &\underline{.961} &\underline{.838} &.891 &\underline{.957} \\
\textsc{SiGAT} & .753&.637 &.658 &.790 &.640 &.752 &.772 &.651 &.725 \\
\textsc{SDGNN} &.813 &.683 &.743 &\underline{.839} &.674 &.825 &.811 &.718 &.829 \\
\textsc{SPMF}    &.838 &.819 &.902 &.836 &.811 &.904 &.810 &.799 &.917 \\
\textsc{SLIM} &.827 &\underline{.897} &.933 &.837 &\textbf{.902} &.947 &.808 &\underline{.895} &.941  \\
\midrule
\textsc{\modelab (ours)}   &\textbf{.863} &\textbf{.898} &\textbf{.969} &\textbf{.867} &\underline{.898} &\textbf{.965} &\textbf{.839} &\textbf{.903} &\textbf{.972}\\
\bottomrule    
\end{tabular}%
}
\end{table*}

\begin{table*}[h]
\centering
\caption{Area Under Curve (AUC-PR) scores for representation size of $K=8$.}
\label{tab:auc_pr}
\resizebox{0.6\textwidth}{!}{%
\begin{tabular}{rcccccccccccccccccccccc}\toprule
\multicolumn{1}{l}{} & \multicolumn{3}{c}{\textsl{Homo sapiens}} & \multicolumn{3}{c}{\textsl{Mus musculus}} & \multicolumn{3}{c}{\textsl{Rattus norvegicus}}\\
\cmidrule(rl){2-4}\cmidrule(rl){5-7}\cmidrule(rl){8-10}
\multicolumn{1}{r}{Task} & $p@n$ & $p@z$ & $n@z$ & $p@n$ & $p@z$ & $n@z$ & $p@n$ & $p@z$ & $n@z$  \\\cmidrule(rl){1-1}\cmidrule(rl){2-2}\cmidrule(rl){3-3}\cmidrule(rl){4-4}\cmidrule(rl){5-5}\cmidrule(rl){6-6}\cmidrule(rl){7-7}\cmidrule(rl){8-8}\cmidrule(rl){9-9}\cmidrule(rl){10-10}
\textsc{POLE}    &.894 &.775 &.518 &.918 &.769 &.586 &.907 &.761 &.577 \\
\textsc{SLF} &\underline{.946} &.868 &\underline{.831} &.949 &.874 &\underline{.851} &\underline{.947} &.869 &.838 \\
\textsc{SiGAT} &.901 &.615 &.338 &.928 &.607 &.400 &.922 &.615 &.349 \\
\textsc{SDGNN} &.938 &.544 &.319 &\underline{.952} &.608 &.443 &.942 &.653 &.547 \\
\textsc{SPMF}    &.944 &.768 &.692 &.950 &.772 &.687 &.939 &.742 &.724 \\
\textsc{SLIM} &.936 &\underline{.889} & \underline{.831} &.947 &\underline{.894} &.824  &.936 &\underline{.887} &\underline{.847} \\
\midrule
\textsc{\modelab (ours)}    &\textbf{.954} &\textbf{.895} &\textbf{.895} &\textbf{.960} &\textbf{.898} &\textbf{.873} &\textbf{.950} &\textbf{.904} &\textbf{.900}\\
\bottomrule    
\end{tabular}%
}
\end{table*}

\subsection*{Training details}
We continue by providing the experimental set-up, the considered datasets and baselines for evaluating the performance and robustness of our proposed framework. All experiments regarding our proposed model and baselines have been conducted on an $8$ GB NVIDIA RTX $2070$ Super GPU. Furthermore, for training, we used the Adam optimizer \citep{kingma2017adam} with a fixed learning rate $\text{lr}=0.05$ and performed $5000$ iterations. Lastly, the proposed \textsc{S$2$-SPM} was initialized based on the furthest-sum algorithm \citep{5589222,SLIM}.

\newpage

\subsection*{Additional Network Visualizations}

\begin{figure*}[h]
\centering
\begin{subfigure}{0.32\textwidth}
    \includegraphics[width=1\textwidth]{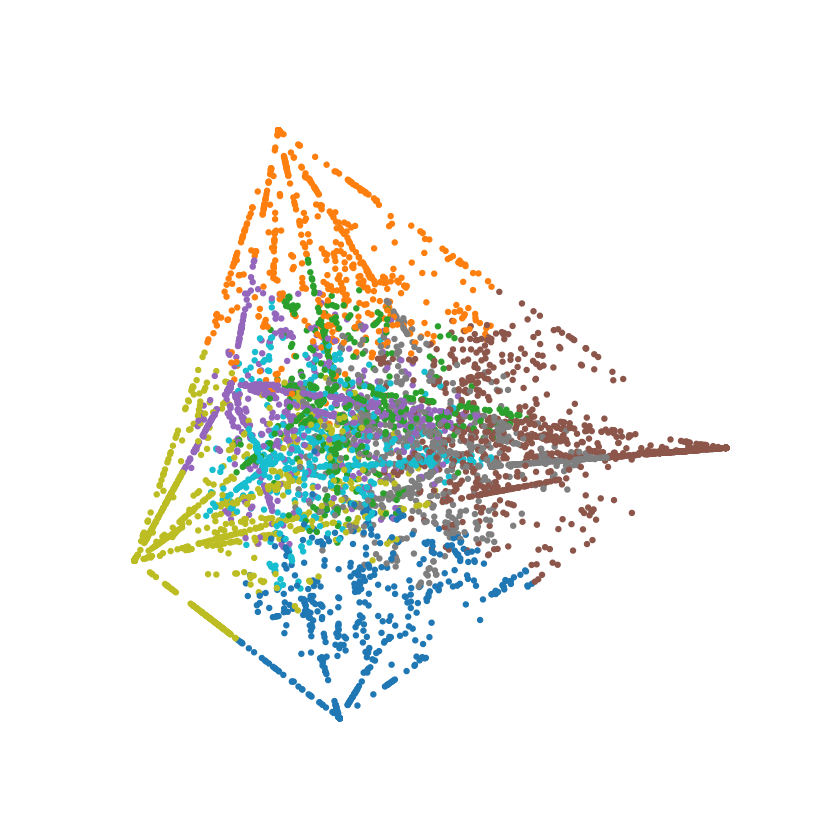}
    \caption{PCA---first two Principal Components}
    
\end{subfigure}
\hfill
\begin{subfigure}{0.32\textwidth}
    \includegraphics[width=1\textwidth]{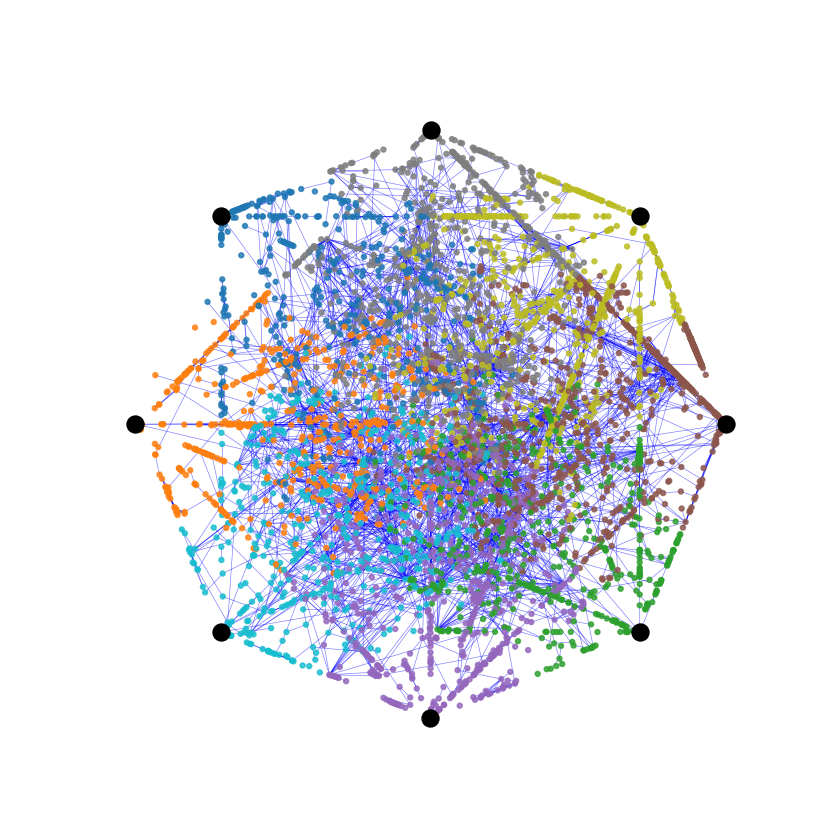}
    \caption{Positive Space Circular Plots}
    
\end{subfigure}
\hfill
\begin{subfigure}{0.32\textwidth}
    \includegraphics[width=0.8\textwidth]{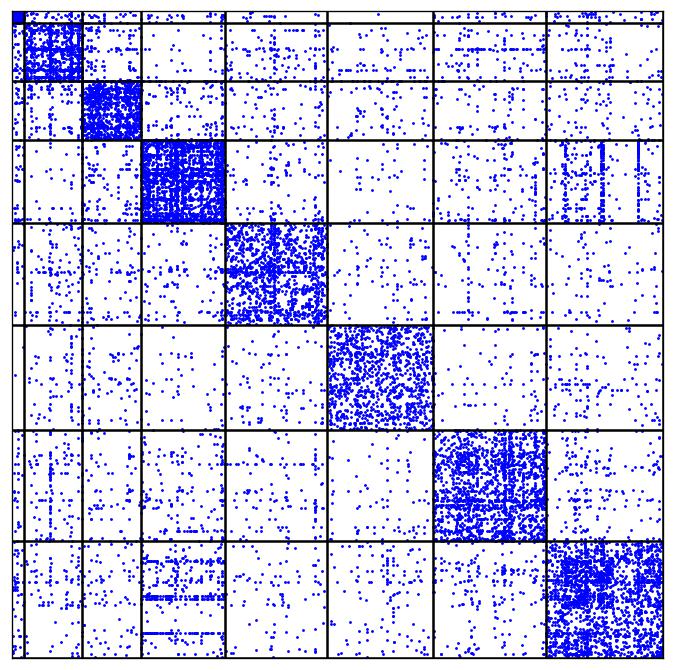}
    \caption{Ordered Positive Adjacency Matrix}
\end{subfigure}

\hfill
\begin{subfigure}{0.32\textwidth}
    \includegraphics[width=1\textwidth]{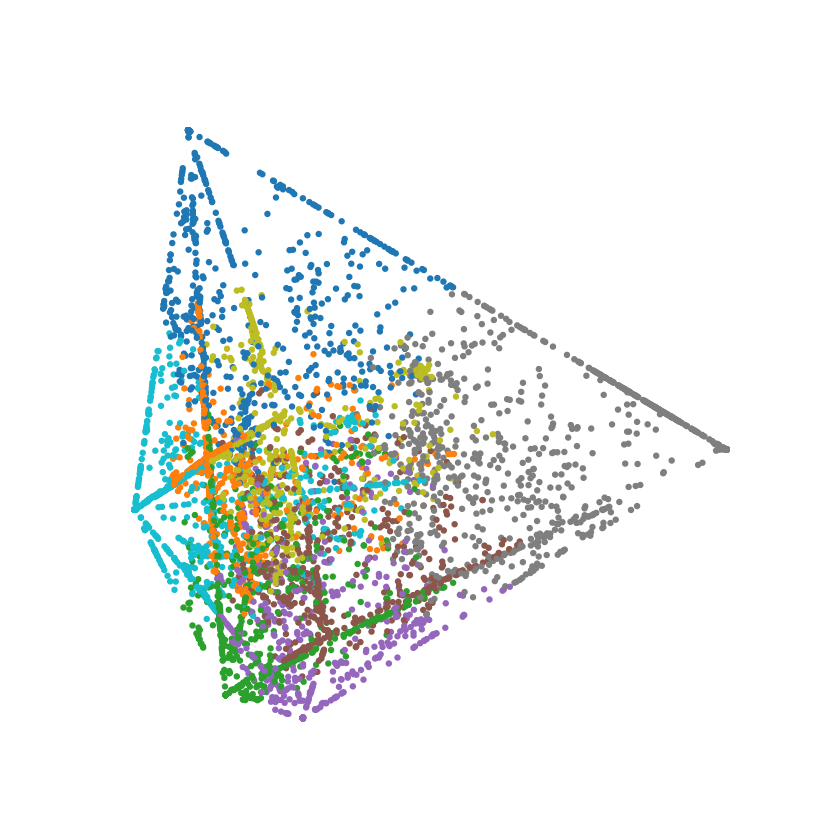}
    \caption{PCA---first two Principal Components}
    
\end{subfigure}
\hfill
\begin{subfigure}{0.32\textwidth}
    \includegraphics[width=1\textwidth]{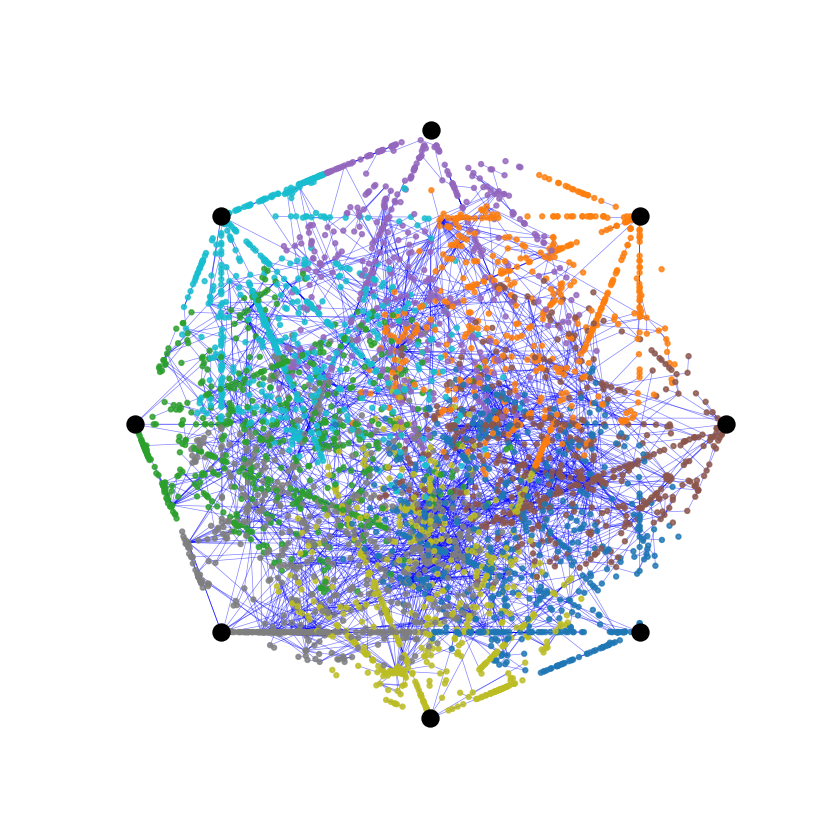}
    \caption{Positive Space Circular Plots}
    
\end{subfigure}
\begin{subfigure}{0.32\textwidth}
    \includegraphics[width=0.8\textwidth]
    {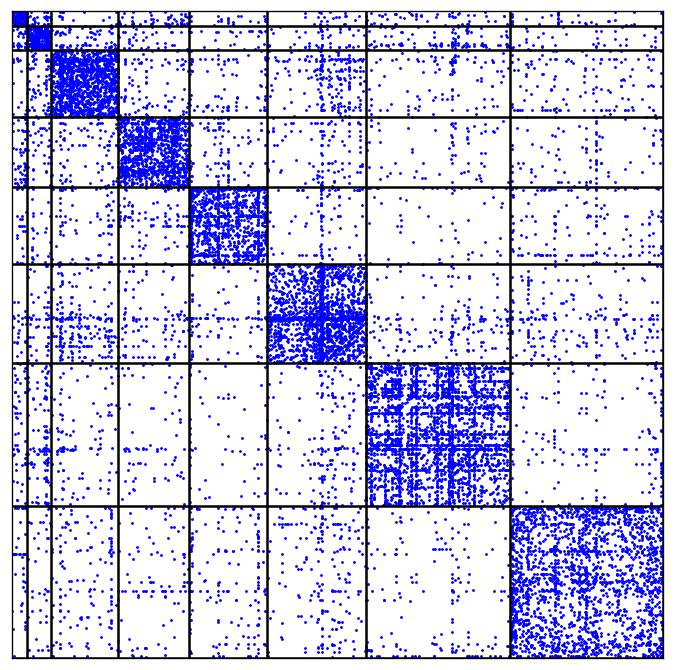}
    \caption{Ordered Positive Adjacency Matrix}
    
\end{subfigure}
\hfill
\begin{subfigure}{0.32\textwidth}
    \includegraphics[width=1\textwidth]{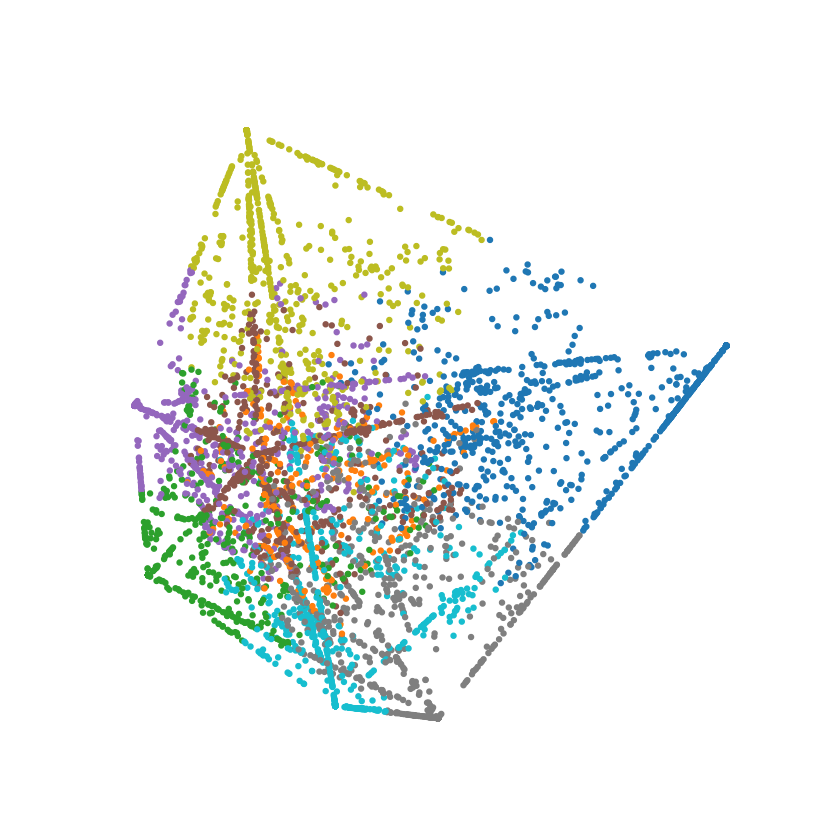}
    \caption{PCA---first two Principal Components}
    
\end{subfigure}
\hfill
\begin{subfigure}{0.32\textwidth}
    \includegraphics[width=1\textwidth]{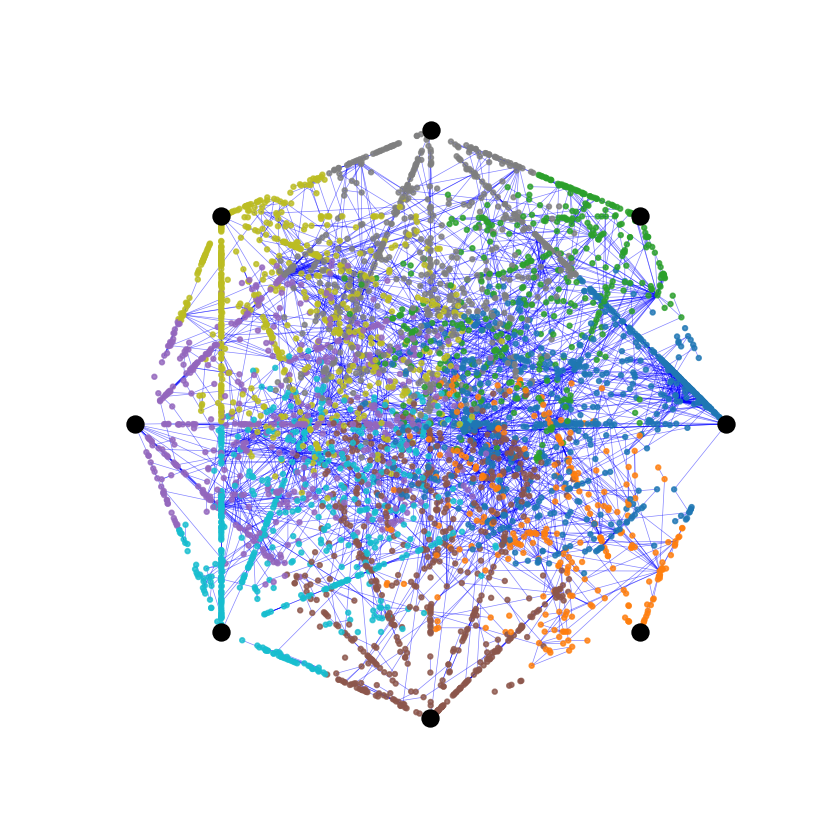}
    \caption{Positive Space Circular Plots}
    
\end{subfigure}
\hfill
\begin{subfigure}{0.32\textwidth}
    \includegraphics[width=0.8\textwidth]{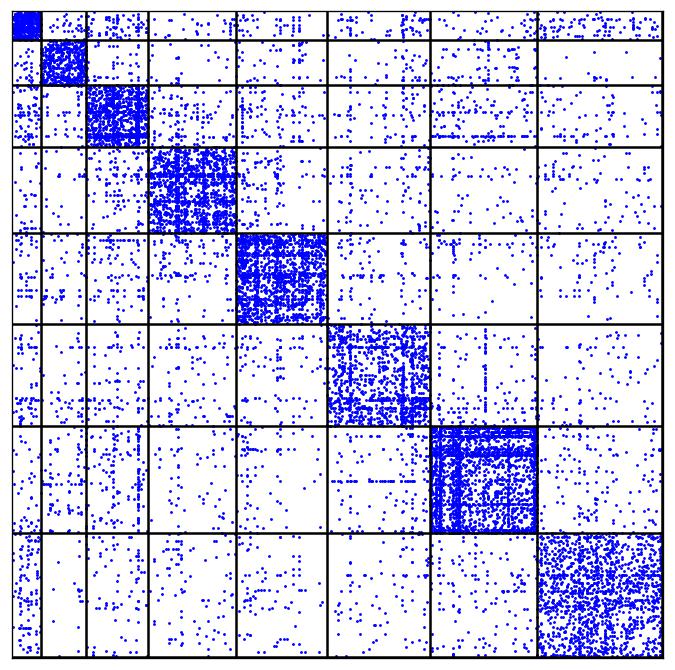}
    \caption{Ordered Positive Adjacency Matrix}
    
\end{subfigure}

\caption{\textbf{\textsc{S$2$-SPM}(K=8)}: Multiple Networks---Positive space inferred simplex visualizations and ordered adjacency matrices for $K=8$ archetypes. The first column shows the latent space projection to the first two Principal Components of the positive latent space $\bm{Z}$---The second column provides the Positive Space Circular Plot (\textsc{PSCP}) with blue lines showcasing positive edges between proteins---The third columns shows the Ordered Positive Edges Adjacency (\textsc{OrA}) matrices sorted based on the memberships $\mathbf{z}_i$, in terms of maximum simplex corner responsibility, and internally according to the magnitude of the corresponding corner assignment for their reconstruction. Each row describes a different dataset, \textsl{Homo sapiens}, \textsl{Mus musculus}, and \textsl{Rattus norvegicus}, respectively.}
\label{fig:polytopes_pos}
\end{figure*}

\begin{figure*}[h]
\centering
\begin{subfigure}{0.32\textwidth}
    \includegraphics[width=1\textwidth]{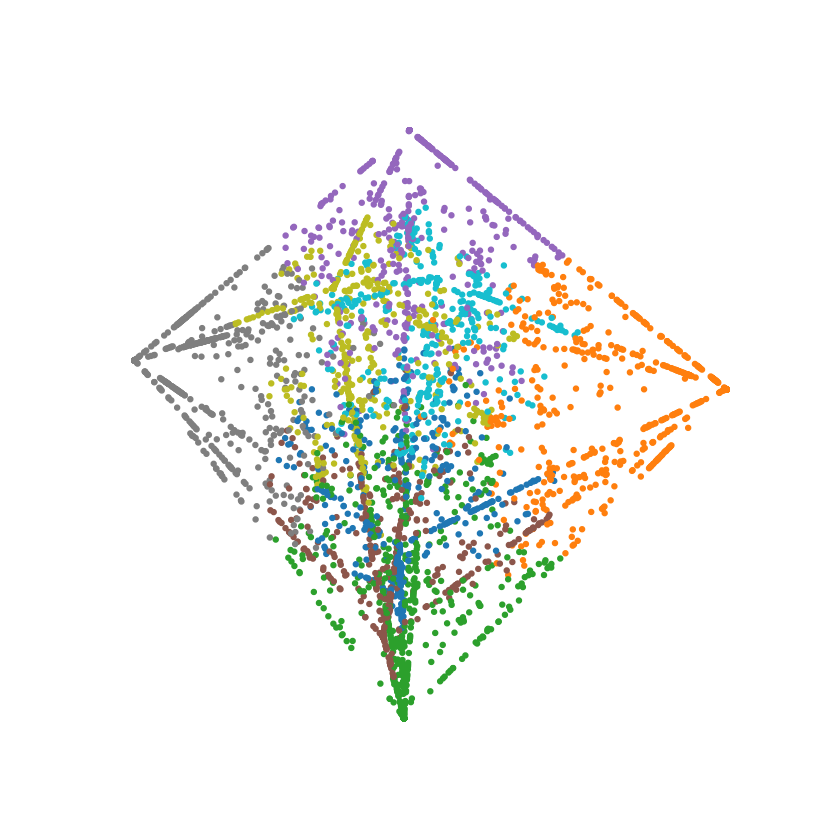}
    \caption{PCA---first two Principal Components}
    
\end{subfigure}
\hfill
\begin{subfigure}{0.32\textwidth}
    \includegraphics[width=1\textwidth]{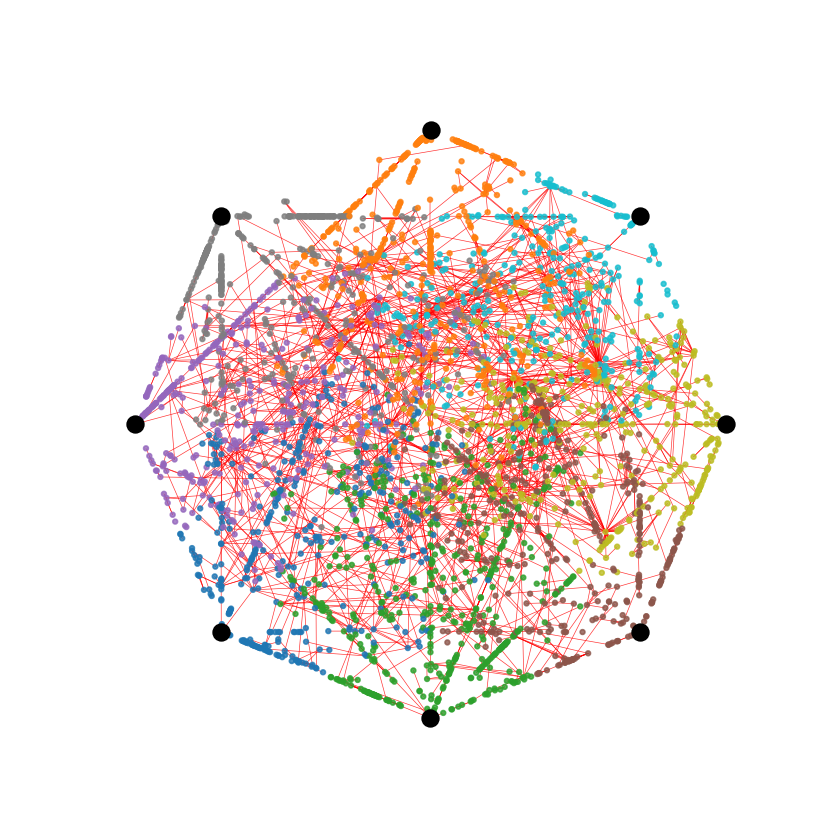}
    \caption{Negative Space Circular Plots}
    
\end{subfigure}
\hfill
\begin{subfigure}{0.32\textwidth}
    \includegraphics[width=0.8\textwidth]{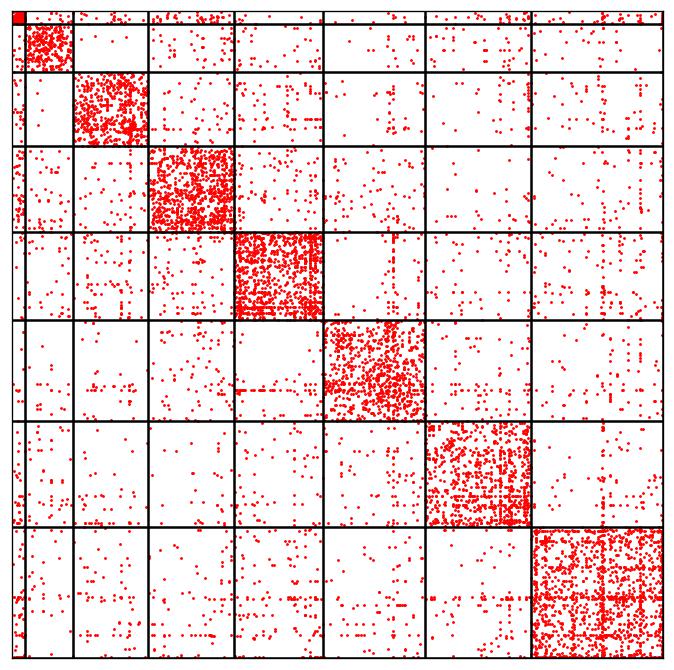}
    \caption{Ordered Negative Adjacency Matrix}
    
\end{subfigure}

\hfill
\begin{subfigure}{0.32\textwidth}
    \includegraphics[width=1\textwidth]{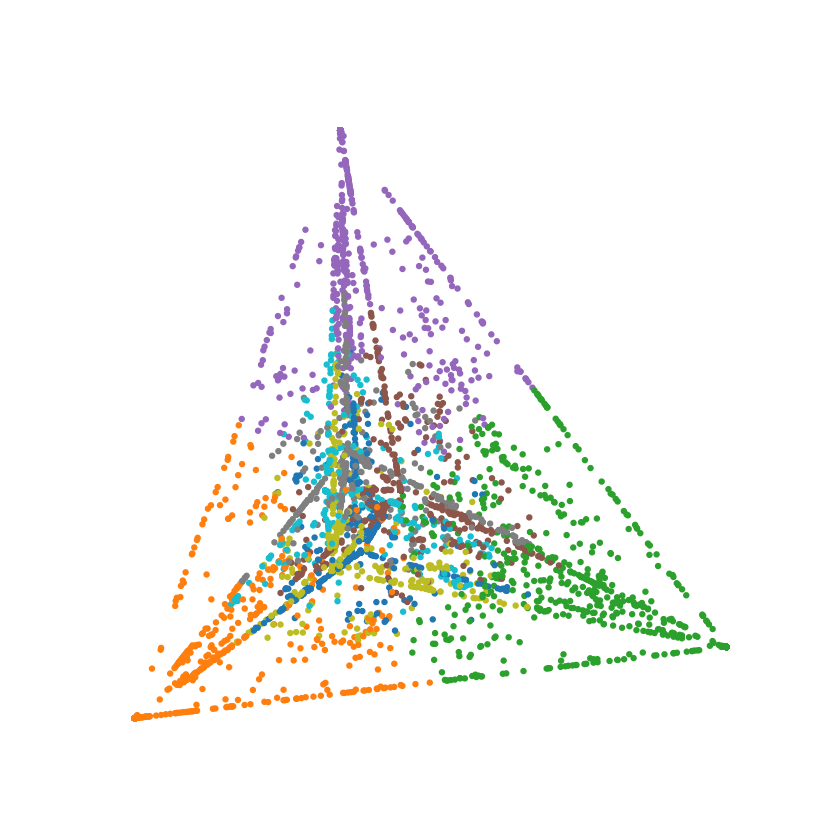}
    \caption{PCA---first two Principal Components}
    
\end{subfigure}
\hfill
\begin{subfigure}{0.32\textwidth}
    \includegraphics[width=1\textwidth]{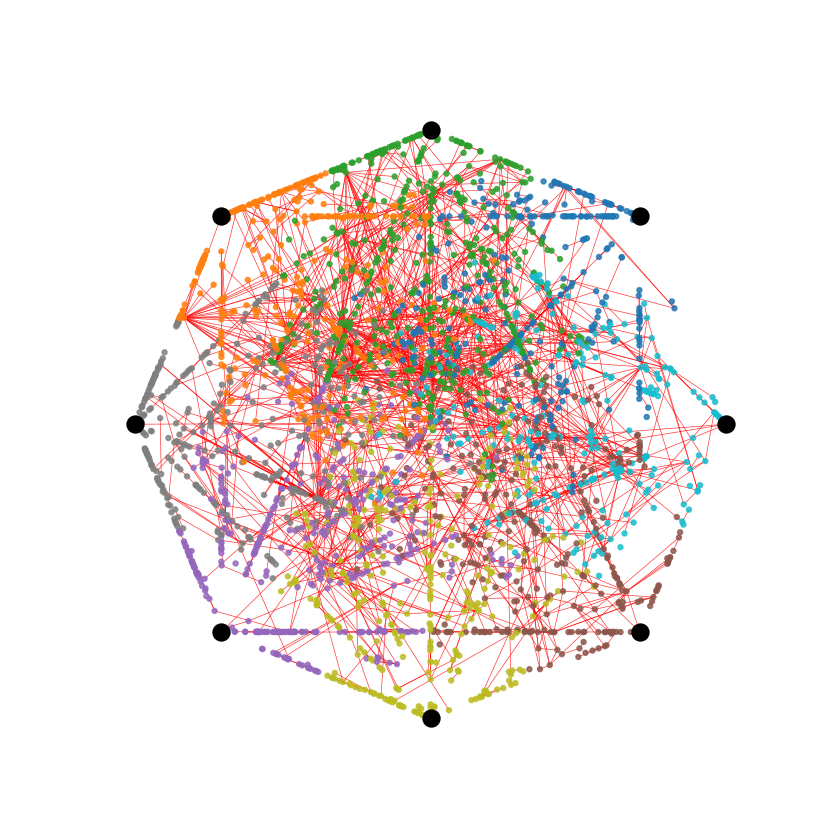}
    \caption{Negative Space Circular Plots}
    
\end{subfigure}
\begin{subfigure}{0.32\textwidth}
    \includegraphics[width=0.8\textwidth]
    {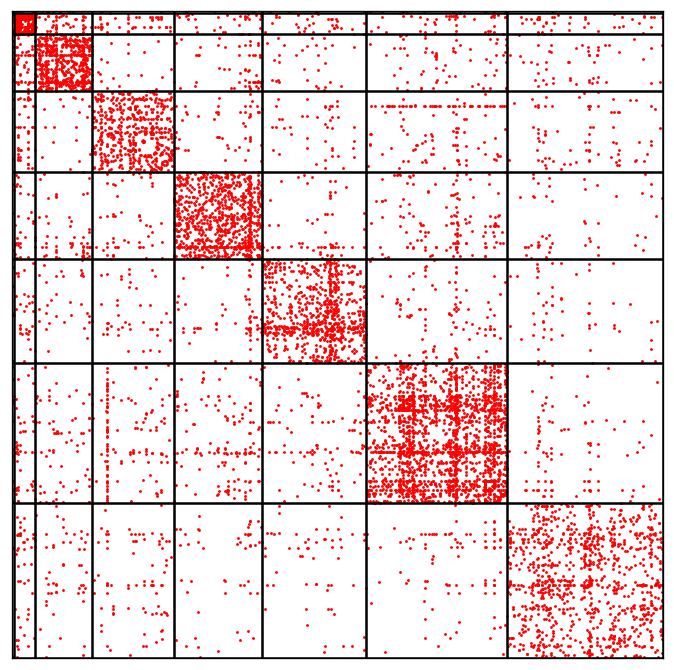}
    \caption{Ordered Negative Adjacency Matrix}
    
\end{subfigure}
\hfill
\begin{subfigure}{0.32\textwidth}
    \includegraphics[width=1\textwidth]{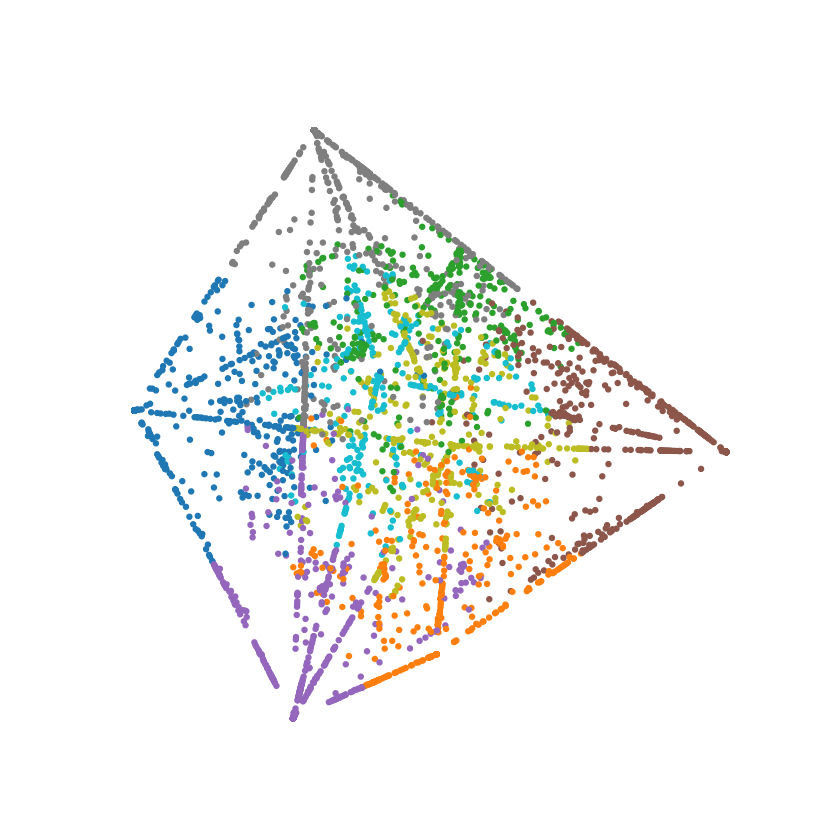}
    \caption{PCA---first two Principal Components}
    
\end{subfigure}
\hfill
\begin{subfigure}{0.32\textwidth}
    \includegraphics[width=1\textwidth]{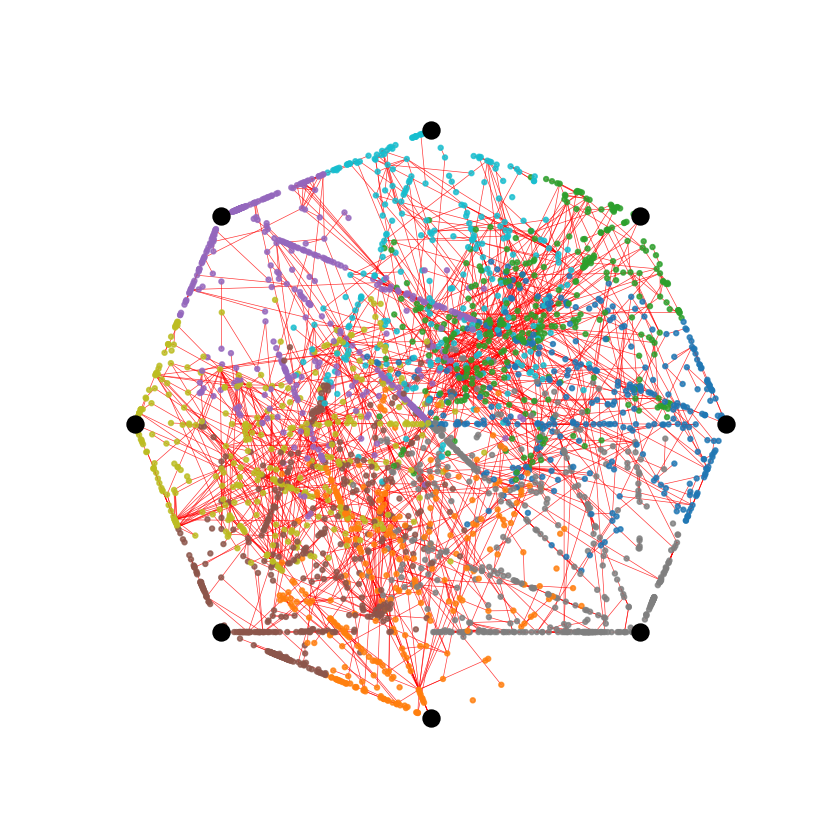}
    \caption{Negative Space Circular Plots}
    
\end{subfigure}
\hfill
\begin{subfigure}{0.32\textwidth}
    \includegraphics[width=0.8\textwidth]{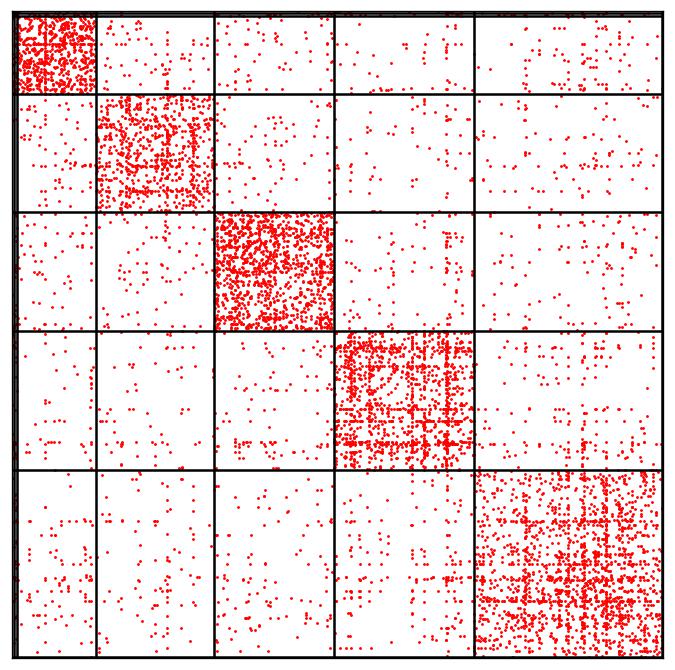}
    \caption{Ordered Negative Adjacency Matrix}
    
\end{subfigure}

\caption{\textbf{\textsc{S$2$-SPM}(K=8)}: Multiple Networks---Negative space inferred simplex visualizations and ordered adjacency matrices for $K=8$ archetypes. The first column shows the latent space projection to the first two Principal Components of the negative latent space $\bm{W}$---The second column provides the Negative Space Circular Plot (\textsc{NSCP}) with red lines showcasing negative edges between proteins---The third columns shows the Ordered Negative Edges Adjacency (\textsc{OrA}) matrices sorted based on the memberships $\mathbf{w}_i$, in terms of maximum simplex corner responsibility, and internally according to the magnitude of the corresponding corner assignment for their reconstruction. Each row describes a different dataset, \textsl{Homo sapiens}, \textsl{Mus musculus}, and \textsl{Rattus norvegicus}, respectively.}
\label{fig:polytopes_neg}
\end{figure*}
\clearpage

\subsection*{Examples of enriched GO terms at each archetype}

\begin{table*}[h]
\centering
\caption{\textsl{Homo sapiens}---Negative space archetype enrichment analysis for representation size of $K=8$.}
\label{tab:enrichement_neg}
\resizebox{1\textwidth}{!}{%
\begin{tabular}{cccc}
\toprule
\textsc{Archetype 1} & \textsc{Archetype 2} & \textsc{Archetype 3} & \textsc{Archetype 4} \\
\cmidrule(r){1-1} \cmidrule(lr){2-2} \cmidrule(lr){3-3} \cmidrule(l){4-4} 
\begin{tabular}[c]{@{}c@{}}

\textit{P:positive regulation of cell migration}\\ \textit{P:cell morphogenesis}\\ \textit{P:positive regulation of epithelial to mesenchymal transition} \\ \textit{ P:extracellular matrix organization} \\ \textit{P:transforming growth factor beta receptor signaling pathway}

\end{tabular} &
\begin{tabular}[c]{@{}c@{}}

\textit{P:G protein-coupled receptor signaling pathway}\\ \textit{P:positive regulation of cytosolic calcium ion concentration}\\ \textit{P:chemical synaptic transmission} \\ \textit{P:cell surface receptor signaling pathway} \\ \textit{P:cell adhesion}

\end{tabular} &
\begin{tabular}[c]{@{}c@{}}

\textit{F:guanyl-nucleotide exchange factor activity}\\ \textit{F:hormone activity}\\ \textit{P:cell-cell signaling}\\ \textit{F:receptor ligand activity} \\ \textit{C:extracellular space}

\end{tabular} &
\begin{tabular}[c]{@{}c@{}}

\textit{F:GTPase activator activity}\\ \textit{P:regulation of small GTPase mediated signal transduction}\\ \textit{P:Rho protein signal transduction} \\ \textit{C:late endosome membrane
}

\end{tabular} \\
\bottomrule    
\toprule
\textsc{Archetype 5} & \textsc{Archetype 6} & \textsc{Archetype 7} & \textsc{Archetype 8} \\
\cmidrule(r){1-1} \cmidrule(lr){2-2} \cmidrule(lr){3-3} \cmidrule(l){4-4} 
\begin{tabular}[c]{@{}c@{}}

\textit{P:defense response to virus}\\ \textit{P:innate immune response}\\ \textit{F:mRNA binding}\\ \textit{P:positive regulation of interferon-alpha production} \\ \textit{ P:positive regulation of type I interferon production}

\end{tabular} &
\begin{tabular}[c]{@{}c@{}}

\textit{C:mitochondrial matrix}\\ \textit{F:tRNA binding} \\ \textit{C:nucleosome}

\end{tabular} &
\begin{tabular}[c]{@{}c@{}}

\textit{C:mitochondrion}\\ \textit{P:positive regulation of autophagy}\\ \textit{P:intrinsic apoptotic signaling pathway in response to DNA damage} \\ \textit{P:positive regulation of intrinsic apoptotic signaling pathway}

\end{tabular} &
\begin{tabular}[c]{@{}c@{}}

\textit{P:cell division}\\ \textit{P:mitotic cell cycle}\\ \textit{P:MAPK cascade} \\ \textit{P:microtubule cytoskeleton organization} \\ \textit{F:myosin phosphatase activity}\end{tabular} \\
\bottomrule    
\end{tabular}%
}
\end{table*}

\begin{table*}[h]
\centering
\caption{\textsl{Homo sapiens}---Positive space archetype enrichment analysis for representation size of $K=8$.}
\label{tab:enrichement_pos}
\resizebox{1\textwidth}{!}{%
\begin{tabular}{cccc}
\toprule
\textsc{Archetype 1} & \textsc{Archetype 2} & \textsc{Archetype 3} & \textsc{Archetype 4} \\
\cmidrule(r){1-1} \cmidrule(lr){2-2} \cmidrule(lr){3-3} \cmidrule(l){4-4} 
\begin{tabular}[c]{@{}c@{}}

\textit{P:osteoblast differentiation}\\ \textit{P:transforming growth factor beta receptor signaling pathway}\\ \textit{P:cell morphogenesis} \\ \textit{P:positive regulation of immune response} \\ \textit{P:regulation of transcription by RNA polymerase II}

\end{tabular} &
\begin{tabular}[c]{@{}c@{}}

\textit{P:microtubule cytoskeleton organization}\\ \textit{P:chromosome segregation}\\ \textit{C:chromosome, centromeric region} \\ \textit{P:DNA repair} 

\end{tabular} &
\begin{tabular}[c]{@{}c@{}}

\textit{P:translation}\\  \textit{P:mitotic cell cycle} \\ \textit{P:modulation of chemical synaptic transmission}

\end{tabular} &
\begin{tabular}[c]{@{}c@{}}

\textit{P:ubiquitin-dependent protein catabolic process}\\ \textit{P:macroautophagy}\\ \textit{P:protein ubiquitination}

\end{tabular} \\
\bottomrule    
\toprule
\textsc{Archetype 5} & \textsc{Archetype 6} & \textsc{Archetype 7} & \textsc{Archetype 8} \\
\cmidrule(r){1-1} \cmidrule(lr){2-2} \cmidrule(lr){3-3} \cmidrule(l){4-4} 
\begin{tabular}[c]{@{}c@{}}

\textit{P:G protein-coupled receptor signaling pathway}\\ \textit{P:cellular response to hormone stimulus} \\ \textit{P:neuropeptide signaling pathway} \\ \textit{P:calcium-mediated signaling
}

\end{tabular} &
\begin{tabular}[c]{@{}c@{}}

\textit{F:structural constituent of chromatin}\\ 
\textit{P:Wnt signaling pathway}

\end{tabular} &
\begin{tabular}[c]{@{}c@{}}

\textit{P:ubiquitin-dependent protein catabolic process}\\ \textit{P:proteasome-mediated ubiquitin-dependent protein catabolic process}

\end{tabular} &
\begin{tabular}[c]{@{}c@{}}

\textit{P:defense response to virus}\\
\textit{P:defense response to bacterium}\\ 
\textit{P:inflammatory response}\\
\textit{P:innate immune response}\\
\textit{P:adaptive immune response}

\end{tabular} \\
\bottomrule    
\end{tabular}%
}
\end{table*}

\bibliographystyle{abbrvnat}

\bibliography{reference}

%% file: intro_new.tex
\section{Introduction}
Proteins interact with each other to carry out various cellular functions \citep{legrain2001protein}, forming complex networks within biological pathways known as the interactome \citep{cusick2005interactome}.
Capturing protein-protein interactions (PPIs) is fundamental for decoding cellular processes, crucially implicated in disease mechanisms \citep{rual2005towards}.
However, experimental methods for the determination of PPIs, such as yeast two-hybrid systems \citep{ito2001comprehensive} and mass spectrometry \citep{gavin2002functional}, are costly, time-consuming and insufficient \citep{han2005effect}.
Recently, machine learning (ML) techniques provide accurate alternatives to overcome these challenges.

ML methods for modeling PPI networks \citep{soleymani2022protein,tang2023machine} fall into two main categories: (i) sequence- and structure-based approaches that extract protein representations directly from a protein’s primary sequence or 3D structure using features such as amino acid composition, motifs, and structural domains and (ii) treats the PPI network as a graph, leveraging its topology with methods such as network embedding and graph convolutional networks to predict interactions. Sequence-based methods range from traditional methods like SVMs \citep{guo2010pred_ppi} to neural networks \citep{zeng2020protein} that capture non-linear relationships. 
Graph-based approaches, built upon graph neural networks (GNNs) \citep{fout2017protein, tang2024anti}, integrate sequence and structural information for improved predictions. Specifically, structure-based methods utilize 3D protein data \citep{liu2020combining}, with recent advancements incorporating hierarchical GNN architectures \citep{gao2023hierarchical}. On the other hand, link prediction techniques, such as L3 \citep{kovacs2019network} and similarity-based methods \citep{yuen2023normalized}, focus on network topology, considering properties like node degrees and community partitions. 
Protein-protein interaction networks can be modeled as signed, capturing both activating (positive) and inhibitory (negative) interactions extending prediction beyond the presence of a link. This enables a more natural representation of biological relationships, essential for modeling complex regulatory mechanisms \citep{pritykin2013simple}.
Several representation methods for signed graphs have been proposed over the years for tasks spanning from community detection \citep{signed_com1} to link prediction \citep{slink1}, mostly for social network applications. 
Representative models have been built upon the psychological-based concept of balance theory \citep{cartwright1956structural} and random walks \citep{perozzi2014deepwalk} to capture interactions, such as \textsc{SIDE} \citep{side} and \textsc{POLE} \citep{pole}.
Extending these ideas with neural networks, \textsc{SiGNET} \citep{signet} combines multi-layer perceptrons with balance theory and \textsc{SLF} \citep{slf} introduces multiple latent factors that model additional types of interactions.
Built upon GNNs, \textsc{SiGAT} \citep{sigat} combines common signed network concepts with graph attention networks.
A more recent approach, \textsc{SPMF} \citep{SPMF}, extracts node representations using low-rank matrix approximation to better encode multi-order signed proximity.

Unsupervised learning and clustering techniques play a key role in uncovering hidden patterns in protein-protein networks \citep{pro_clus}. Archetypal Analysis (AA) \citep{cutler1994a,5589222}, originally developed for analyzing observational data within a 
$K$-dimensional polytope, has been extended to fields like computer vision \citep{Chen_2014_CVPR} and population genetics \citep{Gimbernat-Mayol2021.11.28.470296}. 
More recently, AA has been adapted for relational data \citep{SLIM,SGAAE}, including applications to signed social networks under the Skellam distribution \citep{jg1946frequency}. The \textsc{SLIM} method \citep{SLIM} introduces a unified embedding space, where positive links bring nodes closer in a latent ``sociotope," while negative links push them apart. Although effective for modeling social relationships, this approach does not directly extend to signed protein-protein interaction networks (\textsc{SPPI}) and requires additional modeling adaptations.

In this study, we propose the Signed Two Space Proximity Model (S$2$-SPM), the first archetypal-based signed network specifically tailored to model protein interactions.
The proposed method contributes to the existing research in the field in the following key aspects: (i) S$2$-SPM outperforms all compared baselines in terms of the tasks of sign and signed link prediction
    across three real-world PPI networks by 
    $4.3\%$ on average in F1 score with regards to the best competitor,
    (ii) S$2$-SPM is supported by an enrichment analysis based on the Gene Ontology (GO) terms, clarifying the biological relevance of the identified archetypes. This analysis statistically confirms the validity of the obtained representations and enables potential explainability aspects that can be crucial for the biomedical research community,
    (iii) Extensive visualizations of the obtained latent structures and archetypes highlight the effectiveness of the proposed S$2$-SPM in discovering and capturing latent structures present in \textsc{SPPI} networks,
    (iv) The consistency and robustness of the extracted structures and archetypes are confirmed through informative measures, such as the Bayesian Normalized Mutual Information score. 


%% file: materials_methods.tex
\section{Materials and Methods}

\subsection{Archetypal Analysis}

Archetypes are regarded as the extreme points of the convex hull encompassing the data. Specifically, archetypes refer to the most representative or extreme examples within the dataset, which can be used to understand the essential characteristics or patterns present in the data. They serve as the utmost manifestations of data traits and profiles and can essentially be used to express the data structure in terms of underlying ``archetypal" patterns while facilitating the identification and interpretation of such traits. 

Formally, for a given data matrix $\bm{X}\in\mathbb{R}^{P\times N}$ such as $\bm{X}=\{\bm{x}_1,\bm{x}_2,\dots,\bm{x}_n\}$, we aim to extract the archetype matrix $\bm{A} \in\mathbb{R}^{P\times K}$, $\bm{A}=\{\bm{a}_1,\bm{a}_2,\dots,\bm{a}_K\}$ where $K\ll P$ such that:
\begin{equation}
    \bm{\alpha_j}=\sum_{i=1}^N \bm{x}_i c_{ij},
\end{equation}

\noindent with $\boldsymbol{c}_n\in \Delta^{N-1}$, where $\Delta^{N-1}$ denotes the standard simplex in $N$ dimensions such that $\bm{c}\in\Delta^{N-1}$ requires  $c_i\geq 0$ and $\|\bm{c}\|_1=1$, (\ie $\sum_i\bm{c}_i=1$). Given that $\bm{A}$ is also the convex hull of the data, each point $\bm{x}_i$ can now be reconstructed as:
\begin{equation}
    \bm{x}_i=\sum_{j=1}^{K} \bm{a}_j z_{ji}
\end{equation}

\noindent where $\boldsymbol{z}_k\in \Delta^{K-1}$, $\Delta^{K-1}$ denotes the $K$-dimensional simplex. The matrix $\bm{Z}$ essentially describes how each data point is expressed as the convex combination of the archetypes defined by $\bm{A}$. The previous can be summarized in a matrix form as:
\begin{eqnarray}
    \mathbf{X}\approx \mathbf{XCZ}\quad
    \text{s.t. }\boldsymbol{c}_n\in \Delta^{N-1} \text{ and } \mathbf{z}_j\in \Delta^{K-1}.
\end{eqnarray}

\noindent The archetypes in this formulation are represented by the columns of $\bm{A}=\bm{X}\bm{C}$, defining the corners of the convex hull, represented as the convex combinations of the data. 

\begin{figure*}[!t]
\centering
    \includegraphics[width=0.9\textwidth]{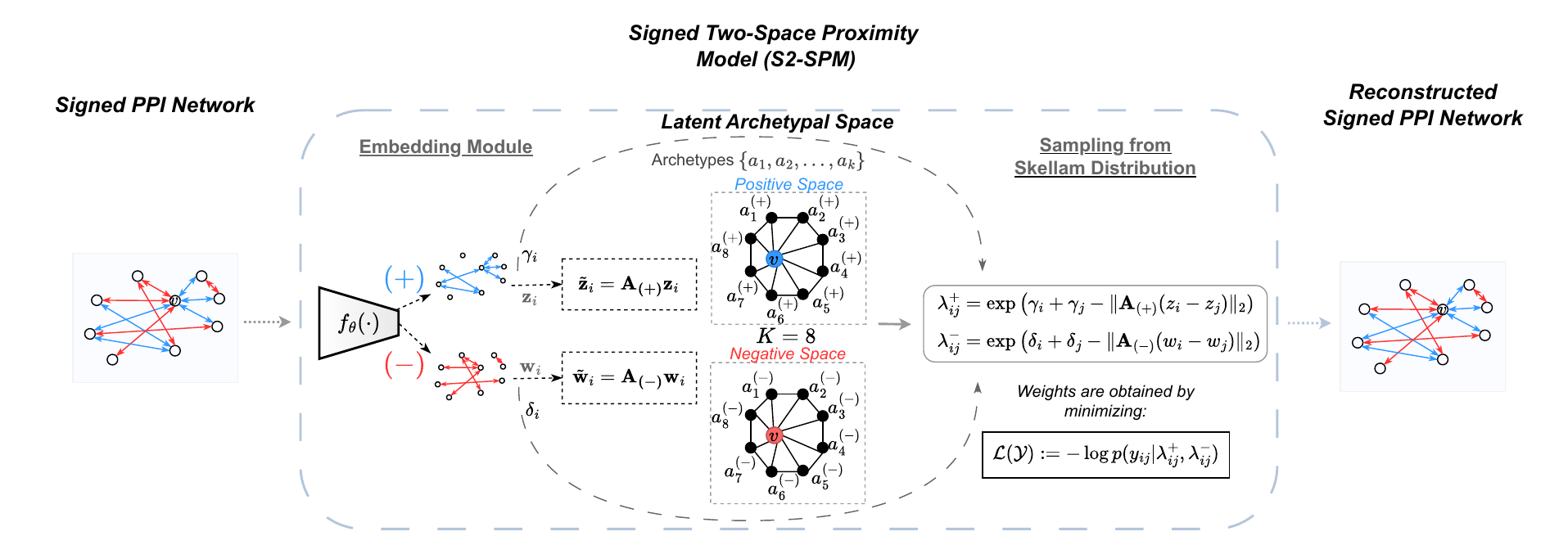}
    \caption{Overview of the proposed Signed Two-Space Proximity Model (\textsc{S$2$-SPM}). Given a signed protein-protein interaction network as an input, the model assigns two latent vectors $\bm{z}_i,\bm{w}_i$ for each of the positive and negative interactions that project each protein to the two archetypal matrices/polytopes $\bm{A}_{(+)}$ and $\bm{A}_{(-)}$, respectively. Then, embeddings are used to calculate the Skellam rates, optimized for the Skellam log-likelihood, to reconstruct the original signed protein-protein graph. } 
    \label{fig:s2spm}
\end{figure*}

\textbf{Archetypal Analysis for Signed Protein-Protein Interaction Networks:} In \textsc{SPPI} networks (see Definition 1), positive and negative interactions should explicitly be accounted for their proximity in the latent space. Specifically, we study \textsc{SPPI} networks where protein interactions take a positive or a negative sign based on the regulation effect (up-regulation/down-regulation). Thus, it is important to analyze and visualize both types of relations, defining nodes as similar and positioning them in close proximity in the latent space. This differentiates from the simplistic assumption in social networks \citep{SLIM} that dissimilar nodes usually interact negatively. Indeed, from a biological point of view, a negative protein-protein interaction should not be translated as animosity between the proteins, as in the case of a ``dislike'' in social networks. 
Thus, in this study, we aim to analyze \textsc{SPPI} networks by differentiating between the signs of interactions in terms of modeling and defining protein similarity independently of the sign of interaction.
\begin{definition}
\it{A signed protein-protein interaction network (SPPI) is a biological network in which nodes represent proteins and edges represent regulatory interactions between them. Each edge is annotated with a sign to specify the regulatory outcome of the interaction: a \textbf{positive $(+)$} edge indicates an \textit{up-regulatory} effect, where the interaction leads to increased activity or expression of the target protein, while a \textbf{negative $(-)$} edge signifies a \textit{down-regulatory} effect, where the interaction results in suppression or inhibition of the target protein.} 
\end{definition}


\textbf{The Signed Two-Space Proximity Model (\textsc{S$2$-SPM}):} 
We next aim to propose a framework for analyzing \textsc{SPPI} networks by projecting them into two independent latent spaces. One space models positive interactions through close proximity, while the other captures negative interactions in a similar fashion. Each latent space is further designed to facilitate archetypal analysis and the characterization of extreme protein profiles within its respective interaction type. This dual-space approach extends previous work \citep{SLIM}—which used a single latent space for social network analysis—by more accurately representing the distinct relational structures inherent to positive and negative interactions. Formally, we aim to learn two sets of latent node representations $\{\mathbf{z}_i\}_{i\in\mathcal{V}}\in\mathbb{R}^{K}$, and $\{\mathbf{w}_i\}_{i\in\mathcal{V}}\in\mathbb{R}^{K}$, defining the two low-dimensional spaces for a given signed network $\mathcal{G}=(\mathcal{V}, \mathcal{Y})$ ($K \ll |\mathcal{V}|$). We assume that the edges of the signed graph can take any integer value representing the intensity of the interaction and the sign (positive/negative) representing the type of connection (up-regulation/down-regulation) of the protein pair. We utilize the Skellam distribution, which is the difference of two independent Poisson-distributed random variables ($y=N_1 - N_2\in\mathbb{Z}$) with respect to the rates $\lambda^{+}$ and $\lambda^{-}$: 
\begin{align*}
P(y|\lambda^{+},\lambda^{-}) = e^{-(\lambda^{+}+\lambda^{-})}\left(\frac{\lambda^{+}}{\lambda^{-}}\right)^{y/2}\mathcal{I}_{|y|}\left(2\sqrt{\lambda^{+}\lambda^{-}}\right),
\end{align*}
where $N_1 \sim Pois(\lambda^{+})$ and $N_2 \sim Pois(\lambda^{-})$, and $\mathcal{I}_{|y|}$ is the modified Bessel function of the first kind and order $|y|$. In general, $\lambda^{+}$ generates the intensity of a positive outcome for $y$ while $\lambda^{-}$ that of a negative outcome. Consequently, we can obtain the negative log-likelihood, which acts as our loss function, as:
\begin{align*}
\mathcal{L}(\mathcal{Y}) &:=-\log p(y_{ij}|\lambda^{+}_{ij},\lambda^{-}_{ij}) \\
&= \sum_{i<j}{(\lambda^{+}_{ij}+\lambda^{-}_{ij})} - \frac{y_{ij}}{2}\log\left(\frac{\lambda^{+}_{ij}}{\lambda^{-}_{ij}}\right)-\log(I_{ij}^{*}),
\end{align*}
where $I_{ij}^{*} := \mathcal{I}_{|y_{ij}|}\left(2\sqrt{\lambda^{+}_{ij}\lambda^{-}_{ij}}\right)$. We, here, do not assume any priors over the parameters of the model, contrary to \cite{SLIM}, as we observed better model performance in the former case. Assuming relational data as input, the Skellam distribution rate parameter $\lambda^{+}_{ij}$ is responsible for modeling the intensity of a positive interaction, whereas $\lambda^{-}_{ij}$ the intensity of a negative interaction, for a node pair $\{i,j\}$. In addition, we aim to constrain the latent spaces into polytopes, defining the convex hull of the latent representations and enabling archetypal characterization. For that, we extend the relational AA formulation in two latent spaces to account for independent archetype extraction in these two latent spaces, each responsible for expressing latent similarity based on positive and negative interactions, respectively. We thus define the Skellam rates as:
\begin{align}
    \lambda_{ij}^{+} &=\exp \big( \gamma_{i} + \gamma_{j} - \|\mathbf{A}_{(+)} (\mathbf{z}_{i}-\mathbf{z}_{j})\|_{2}\big)
    \\ 
    &=\exp \big( \gamma_{i} + \gamma_{j} -\|\mathbf{R}_{(+)}\mathbf{Z}\mathbf{C}_{(+)}(\mathbf{z}_{i}-\mathbf{z}_{j})\|_2\big).\label{LRPM_inensity_function_1}
    \\
\lambda_{ij}^{-} &=\exp \big( \delta_{i} + \delta_{j} - \|\mathbf{A}_{(-)} (\mathbf{w}_{i}-\mathbf{w}_{j})\|_{2}\big)
\\ 
&=\exp \big( \delta_{i} + \delta_{j} -\|\mathbf{R}_{(-)}\mathbf{W}\mathbf{C}_{(-)}(\mathbf{w}_{i}-\mathbf{w}_{j})\|_2\big).\label{LRPM_inensity_function_2}
\end{align}

\noindent where $\bm{A}_{(+)}$, and $\bm{A}_{(-)}$ are the matrices with columns containing the archetypes for the positive and negative latent spaces, respectively. This case differs from the classical AA since the data matrices now refer to latent variables. We define the data matrices as $\bm{X}_{(+)}=\bm{R}_{(+)}\bm{Z}$, and $\bm{X}_{(-)}=\bm{R}_{(-)}\bm{W}$ with $\bm{R}_{(+)},\bm{R}_{(-)} \in \mathbb{R}^{K\times K}$. Lastly, to guarantee that the archetypes are points belonging to the latent embeddings, we adopt for $\bm{C}_{(+)},\bm{C}_{(-)}\in \mathbb{R}^{N\times K}$ a gated version \citep{SLIM} as: 
\begin{equation*}
    \begin{aligned}
    c_{(+)nd} &= \frac{(\mathbf{Z}^\top\circ [\sigma(\mathbf{G}_{(+)})]^\top)_{nd}}{\sum_{n^\prime}(\mathbf{Z}^\top\circ [\sigma(\mathbf{G}_{(+)})]^\top)_{n^\prime d}}, \\
    c_{(-)nd} &= \frac{(\mathbf{W}^\top\circ [\sigma(\mathbf{G}_{(-)})]^\top)_{nd}}{\sum_{n^\prime}(\mathbf{W}^\top\circ [\sigma(\mathbf{G}_{(-)})]^\top)_{n^\prime d}},
    \end{aligned}
\end{equation*}

\noindent with $\sigma(\cdot)$ the logistic sigmoid function. Finally, $\{\gamma_i,\delta_i\}_{i\in\mathcal{V}}$ denote the node-specific random effect terms, and $||\cdot||_2$ is the Euclidean distance function. Essentially, $\gamma_i,\gamma_j$ represent the tendency of a node to form positive connections while
$\delta_i,\delta_j$ the tendency to form negative connections. In other words, it accounts for degree heterogeneity in the positive and negative sub-networks, accordingly.

The most closely related work to our method, \textsc{SLIM} \citep{SLIM}, employs a latent distance-based approach to extract a unified embedding space using a Skellam likelihood. However, \textsc{SLIM} does not distinguish between two latent spaces and fails to model negative interactions as close proximity, a crucial feature for SPPIs. Similarly, recent work \citep{SGAAE} introduced two latent membership vectors for positive and negative links but still projects the embeddings onto a shared latent space. In contrast, our method generalizes this approach by decoupling the positive and negative link characterizations into two completely independent polytopes, one for each interaction type.

\textbf{Structure Retrieval and Consistency of Archetypes:} Our proposed approach defines two latent spaces, which are designed to extract archetypes based on the positive link and negative link structures, respectively. \textsl{Protein-protein} networks are known to contain noise and spurious protein interactions \citep{noisy_ppi3,noisy_ppi2,noisy_ppi1}. Consequently, we must verify our models' ability to extract informative, consistent, and robust structures for such networks. For that, we make use of the Bayesian Normalized Mutual Information (BNMI) metric \citep{AA_fmri}, as shown in Eq. \eqref{eq:BNMI}. This is a direct consequence of our model defining soft memberships over the archetypes, and thus, classical clustering quality metrics that are defined over (sole) hard memberships are not optimal. Specifically, we follow \citep{AA_fmri}, and we run our model five times ($r=5$) for each latent dimension $K$ of the model. We then concatenate the obtained positive and negative archetypes membership matrices $\bm{Z}=[\bm{Z}^{(1)};\bm{Z}^{(2)};\bm{Z}^{(3)};\bm{Z}^{(4)};\bm{Z}^{(5)}]$, and $\bm{W}=[\bm{W}^{(1)};\bm{W}^{(2)};\bm{W}^{(3)};\bm{W}^{(4)};\bm{W}^{(5)}]$, respectively. For each case and specification, we calculate the all-pairs \textsc{BNMI} table across the five runs and report the average score and standard deviation based on:

\begin{equation}
    \label{eq:BNMI}
    \textsc{BNMI}(\bm{Q^{r}},\bm{Q^{r'}})=\frac{2\mathcal{I}(\bm{Q^{r}},\bm{Q^{r'}})}{\mathcal{I}(\bm{Q^{r}},\bm{Q^{r}})+\mathcal{I}(\bm{Q^{r'}},\bm{Q^{r'}})},
\end{equation}

\noindent with $\mathcal{I}(\bm{Z_{(+)}^{r}},\bm{Z_{(+)}^{r'}})=\sum_{d,d'}p(d,d')\log \frac{p(d,d')}{p(d)p(d')}$, $p(d,d')=\sum_n p(d/n)p(d'/n)(p(n))$ the joint distribution with $p(d/n)=q_{dn}^r$ while $p(n)=1/N$, and $\mathbf{Q}^{r}=\{\mathbf{Z}^r\}$ or $\mathbf{Q}^{r}=\{\mathbf{W}^{r}\}$. We here note that \textsc{BNMI} is equal to $1$ when $\mathbf{Q}^r=\mathbf{Q}^{r'}$, and equal to $0$ in the case that no common structure/information exists between the two solutions. We also account for structure retrieval sharing due to randomness and calculate the $\textsc{BNMI}$ while randomly permuting the columns of the obtained positive and negative space obtained memberships $\bm{Z}^r$, and $\bm{W}^r$ for each run $r$. Permuting the allocation matrix is expected to destroy the structure, and any shared information will be due to randomness. This procedure allows us to compare the results of our proposed method with the signal one would expect from random noise, providing a metric for their validity.

\subsection{Experimental Setting and Evaluation Protocols}

We continue by providing the experimental set-up and setting, the considered datasets and baselines for evaluating the performance and robustness of our proposed framework.

\textbf{Datasets:} We evaluate the proposed method on \textsl{protein-protein} interaction networks from the SIGnaling Network Open Resource $3.0$ (\textsc{SIGNOR}) \citep{SIGNOR}. Following this line of work, we construct three signed \textsl{protein-protein} interaction networks describing three different organisms; 1) \textsl{Homo sapiens}, 2) \textsl{Mus musculus}, and 3) \textsl{Rattus norvegicus}. To infer these networks we extract all \textsl{protein-protein} pairs in the dataset and assign a positive ($+$) link if the effect describes \textsl{up-regulation} and a negative $(-)$ link for \textsl{down-regulation}. Finally, we extract the largest connected component for each network. Statistics on the three derived networks are provided in the supplementary. 

\textbf{Baselines:} We compare the performance of our model to various state-of-the-art baselines for modeling signed networks. Specifically, (\textbf{i}) \textsc{SLIM} \citep{SLIM} is a latent distance model that learns a single embedding matrix optimizing the Skellam likelihood, (\textbf{ii}) \textsc{POLE} \citep{pole} learns the network embeddings by decomposing the signed random walks auto-covariance similarity matrix, (\textbf{iii}) \textsc{SLF} \citep{slf} extracts representations as the concatenation of two latent factors targeting positive and negative relations, (\textbf{iv}) \textsc{SiGAT} \citep{sigat} is a graph neural network approach that uses graph attention to update the node embeddings, 
(\textbf{v}) \textsc{SDGNN} \citep{SDGNN} combines status and balance theory with a graph neural network to reconstruct link signs, link directions, and signed directed triangles via the node embeddings and (\textbf{vi}) \textsc{SPMF} \citep{SPMF} uses a low-rank matrix approximation to encode the multi-order signed proximity over a signed network yielding expressive node representations.



\textbf{Protein Gene Ontology Terms:} For the proteins included in the datasets, we use the \textsc{UniProt} \citep{uniprot} database to extract the Gene Ontology (GO) terms associated with them. These terms belong to three general categories, including i) \textit{Biological Processes}, ii) \textit{Molecular Functions}, iii) \textit{Cellular Components}.
Biological Process refers to the biological objectives to which the gene or gene product contributes, Molecular Function describes the elemental activities of a gene product at the molecular level, such as binding or catalysis, and Cellular Component denotes the parts of a cell or its extracellular environment where the gene product is active. These annotations are used extensively in the biological sciences for various purposes, including interpreting gene expression patterns and protein-protein interactions. The structured vocabulary allows researchers to make meaningful inferences about protein function based on their GO annotations.
In our study, we focus on analyzing how specific GO terms are represented in the different interaction types—positive and negative—within our networks. By integrating GO annotations, we can attribute functional characteristics to clusters of proteins that frequently interact either positively or negatively, potentially identifying biological pathways or processes that are predominantly regulated by these interaction types.

\textbf{Enrichment Analysis of Archetypes:} Here, we continue with the enrichment analysis of the obtained archetypes from the proposed \textsc{S$2$-SPM}. We consider the model specification defining eight archetypes $(K=8)$ since it provides $BNMI \approx 0.8$ for all three datasets, but the analysis can easily be extended to additional dimensions. In addition, the dimensionality of the two spaces, as introduced by \textsc{S$2$-SPM}, is not required to be the same, \ie  $K_{(+)}\neq K_{(-)}$, we consider though, the case where $K_{(+)}= K_{(-)}=K$, for simplicity. Our enrichment analysis, is based on the protein GO terms while we follow a similar strategy as in \citep{AA_nmeth}, to verify the statistical significance and validity of such an analysis. 
To take into account a particular $GO$ term, at least $20$ proteins need to be related to that term.
For the enrichment analysis of archetype $k$, we start by calculating and sorting the latent distance between every node in the network and the specific archetype. We then define a total of $B$ bins, such that each bin contains an equal amount of network nodes, sorted by distance in increasing order. Consequently, the first bin contains the nodes that reside closest to the archetype, while the last bin contains the points that reside furthest from the archetype. We then search for the GO terms/labels that are maximally enriched in the bin closest to the archetype \citep{AA_nmeth}. We define the enrichment value at a bin $b$ and for a GO term $l$ as $E_{bl}=\rho_{bl}/P_{l}$, where $\rho_{bl}$ is the density of GO term $l$ in the bin $b$, and $P_{l}$ the density of GO term $l$ in the whole dataset. 
As in \citep{AA_nmeth}, we compute the significance of the enrichment value in the bin closest to the archetype $E_{b_0l}$) via a hypergeometric test. Consequently, each GO term is associated with a p-value, describing the significance of the enrichment in the first bin. We consider GO terms with a p-value less than $0.002$ as potential candidates for being enriched in the archetype. Furthermore, to account for the false discovery rate (FDR), given by the high number of performed enrichment significance tests, we perform a multiple-hypothesis test using the Benjamini-Hochberg (BH) procedure, setting the false discovery rate level $a=0.05$. Lastly, for each GO term that is significant under the hypergeometric test ($\text{p-value}<0.002$) and survives the FDR procedure we calculate the probability $p_{\text{max}}$ \citep{AA_nmeth} that the enrichment value in the first bin $E_{b_0l}$ is maximum with regards to the rest of the bins, $E_{b_0l}>E_{bl} \text{ for } b>0$. We finally consider a GO term as enriched in a given archetype if it is associated with a probability $p_{\text{max}}>0.5$. 

\textbf{Bin Size Calculation:} An important aspect of the enrichment analysis adopted in this study is the value of the bin size, which defines the number of points in each bin. In \citep{AA_nmeth}, the authors express the minimum bin size, such that randomness does not affect the underlying enrichment signal. We here argue that a unique choice for the bin size may not be optimal, as it can lead to a different number of enriched GO terms. Based on that, we consider multiple bin sizes ranging from $1\%$ to $20\%$ of the network size $N$ with a step size equal to $1\%$. We consider the final enriched labels as the ones that are characterized as significant (based on the analysis described above) in at least half of the considered bin sizes, \ie yielding a significance appearance rate $\{SAR=\frac{\text{\# bins sizes label is significant}}{\text{\# total bin sizes}}\geq 0.5\}$. By aggregating results based on multiple bin sizes, we argue that any strong dependencies between the enrichment analysis and the choice of the bin size value are removed.

\begin{figure*}[t!]
\centering
\begin{subfigure}{0.24\textwidth}
    \includegraphics[width=\textwidth]{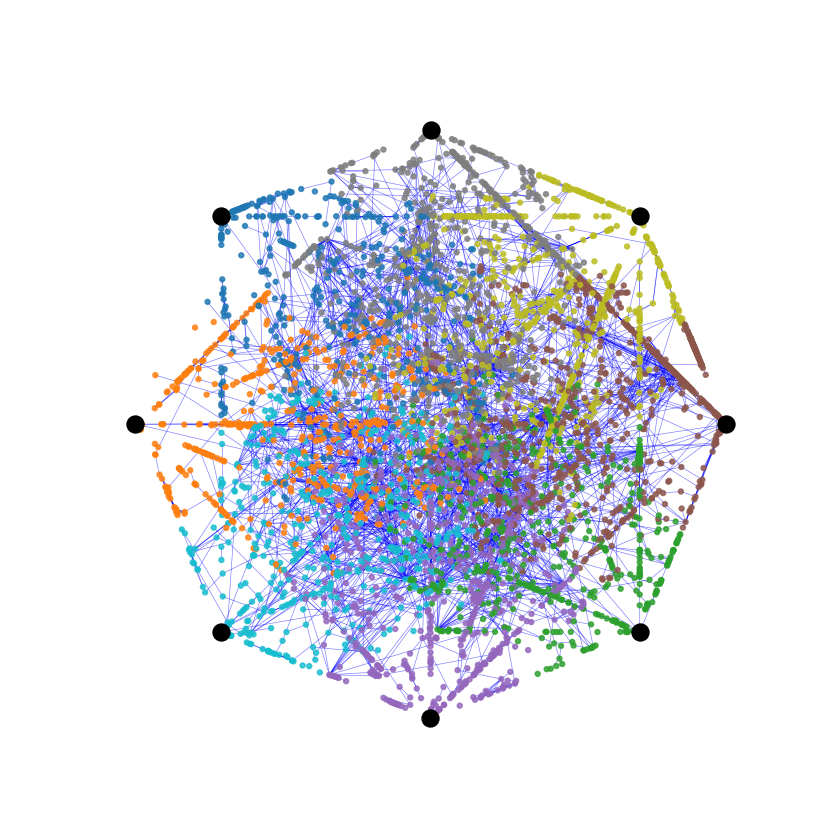}
    \caption{Positive Space Circular Plots}
\end{subfigure}
\hfill
\begin{subfigure}{0.24\textwidth}
    \includegraphics[width=\textwidth]{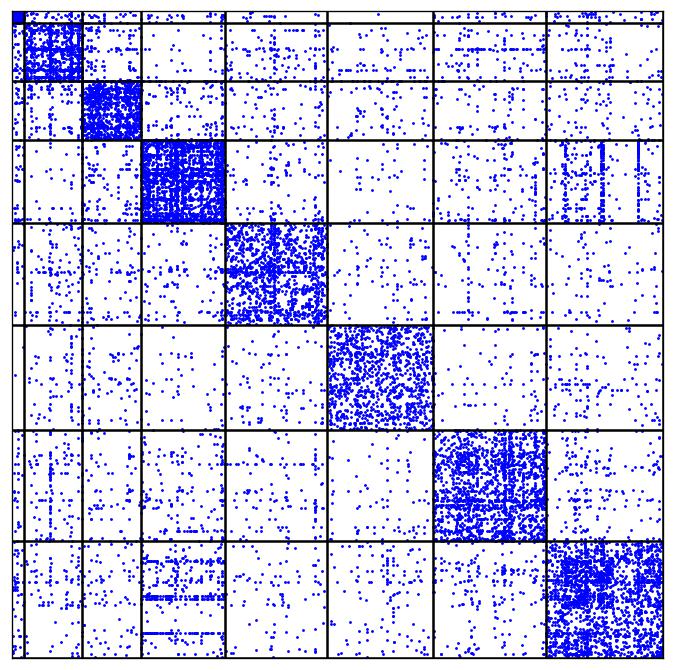}
    \caption{Positive Adjacency Matrix}
\end{subfigure}
\hfill
\begin{subfigure}{0.24\textwidth}
    \includegraphics[width=\textwidth]{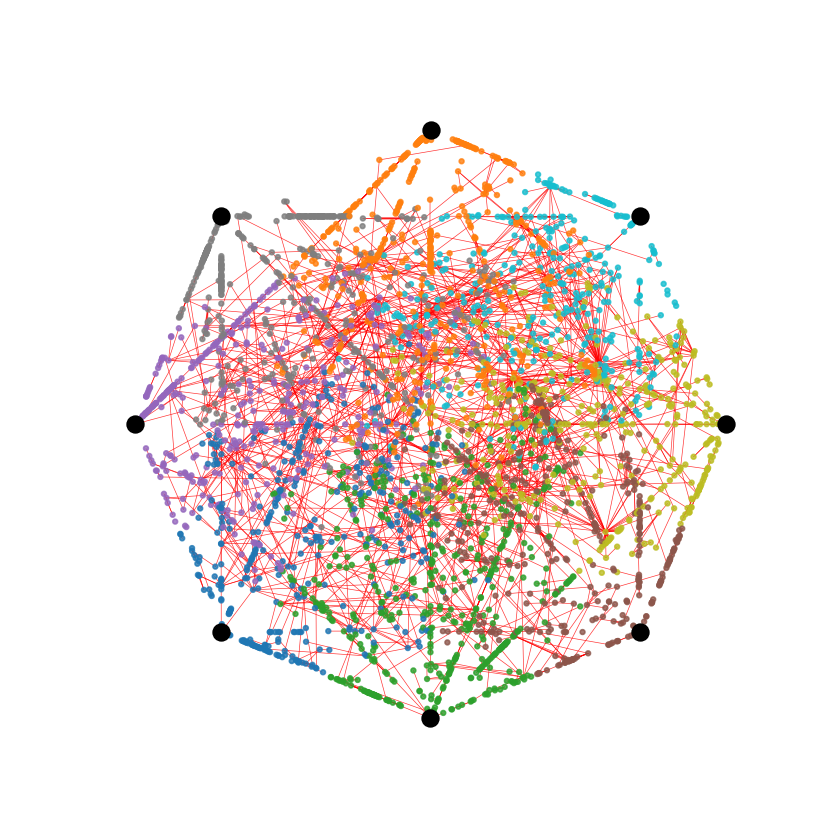}
    \caption{Negative Space Circular Plots}
\end{subfigure}
\hfill
\begin{subfigure}{0.24\textwidth}
    \includegraphics[width=\textwidth]{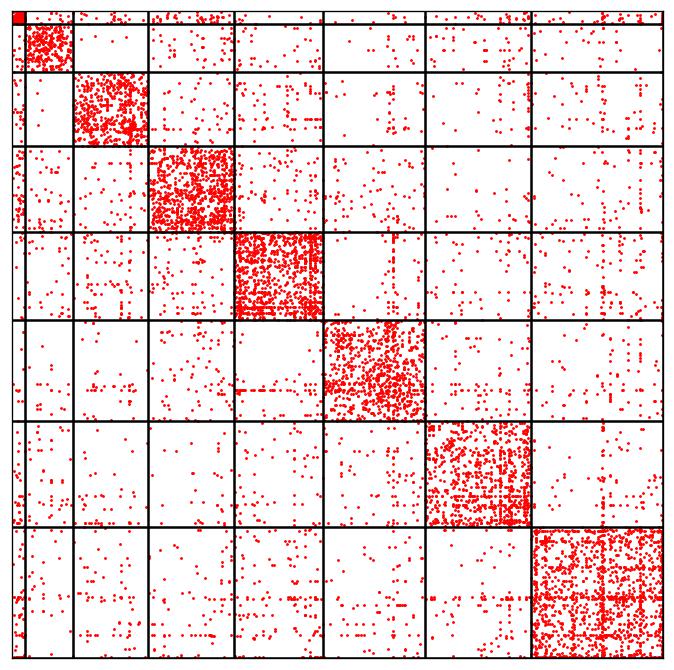}
    \caption{Negative Adjacency Matrix}
\end{subfigure}

\caption{\textbf{\textsl{Homo sapiens} \textsc{S$2$-SPM}(K=8)}: Positive space (a) and (b), and negative space (c) and (d) inferred simplex visualizations and ordered adjacency matrices for $K=8$ archetypes. Figures \ref{fig:polytopes_pos} (a) and (c), provide the Positive/Negative Space Circular Plot (\textsc{PSCP})/(\textsc{NSCP}) with blue/red lines showcasing positive/negative edges between proteins---Figures\ref{fig:polytopes_pos} (b) and (d), show the Ordered Positive/Negative Edges Adjacency (\textsc{OrA}) matrices sorted based on the memberships $\mathbf{z}_i$/$\mathbf{w}_i$, in terms of maximum simplex corner responsibility.}
\label{fig:polytopes_pos}
\end{figure*}

%% file: results.tex
\vspace{-0.2cm}
\section{Results}

\begin{table}[!t]
\centering
\caption{Per class ($pos$, $zr$, $neg$) and weighted ($w$) F1 Scores for representation size $K=8$.}
\label{tab:f1_score}
\resizebox{0.48\textwidth}{!}{%
\begin{tabular}{rcccccccccccccccccccccc}\toprule
\multicolumn{1}{l}{} & \multicolumn{4}{c}{\textsl{Homo sapiens}} & \multicolumn{4}{c}{\textsl{Mus musculus}} & \multicolumn{4}{c}{\textsl{Rattus norvegicus}}\\
\cmidrule(rl){2-5}\cmidrule(rl){6-9}\cmidrule(rl){10-13}
\multicolumn{1}{r}{Class} & $neg$ & $zr$ & $pos$ & $w$ & $neg$ & $zr$ & $pos$ & $w$ & $neg$ & $zr$ & $pos$ & $w$ \\\cmidrule(rl){1-1}\cmidrule(rl){2-2}\cmidrule(rl){3-3}\cmidrule(rl){4-4}\cmidrule(rl){5-5}\cmidrule(rl){6-6}\cmidrule(rl){7-7}\cmidrule(rl){8-8}\cmidrule(rl){9-9}\cmidrule(rl){10-10}\cmidrule(rl){11-11}\cmidrule(rl){12-12}\cmidrule(rl){13-13}
\textsc{POLE}    &.292 &.738 &.413 &.563 &.317 & .746& .449&.585 &.373 &.758 &.465 &.603 \\
\textsc{SLF} &\underline{.525} &\underline{.832} &\underline{.683} &\underline{.740} &\underline{.526} &\underline{.826} &\underline{.661} &\underline{.729} &\underline{.504} &.823 &.670 &.729 \\
\textsc{SiGAT} &.278 &.691 &.513 &.575 &.322 &.679 &.503 &.572 &.375 &.713 &.563 &.618 \\
\textsc{SDGNN} &.447 &.735 &.565 &.637 &.459 &.729 &.557 &.633 &.435 &.721 & .564&.629 \\
\textsc{SPMF}    &.375 &.689 &.558 &.602 &.344 &.680 &.549 &.593 &.318 &.687 &.582 &.606 \\
\textsc{SLIM} &.463 &.826 &.649 &.716 &.445 &.822 &.652 &.715 &\textbf{.506} &\underline{.836} &\underline{.682} &\underline{.740} \\
\midrule 
\textsc{\modelab}    &\textbf{.562} &\textbf{.852} &\textbf{.704} &\textbf{.761}&\textbf{.541} &\textbf{.851} &\textbf{.702} &\textbf{.759} &\underline{.504} &\textbf{.863} &\textbf{.706} &\textbf{.763} \\
\bottomrule    
\end{tabular}%
}
\end{table}

\begin{figure}[!htb]
\centering
\begin{subfigure}{0.49\columnwidth}
    \includegraphics[width=\textwidth]{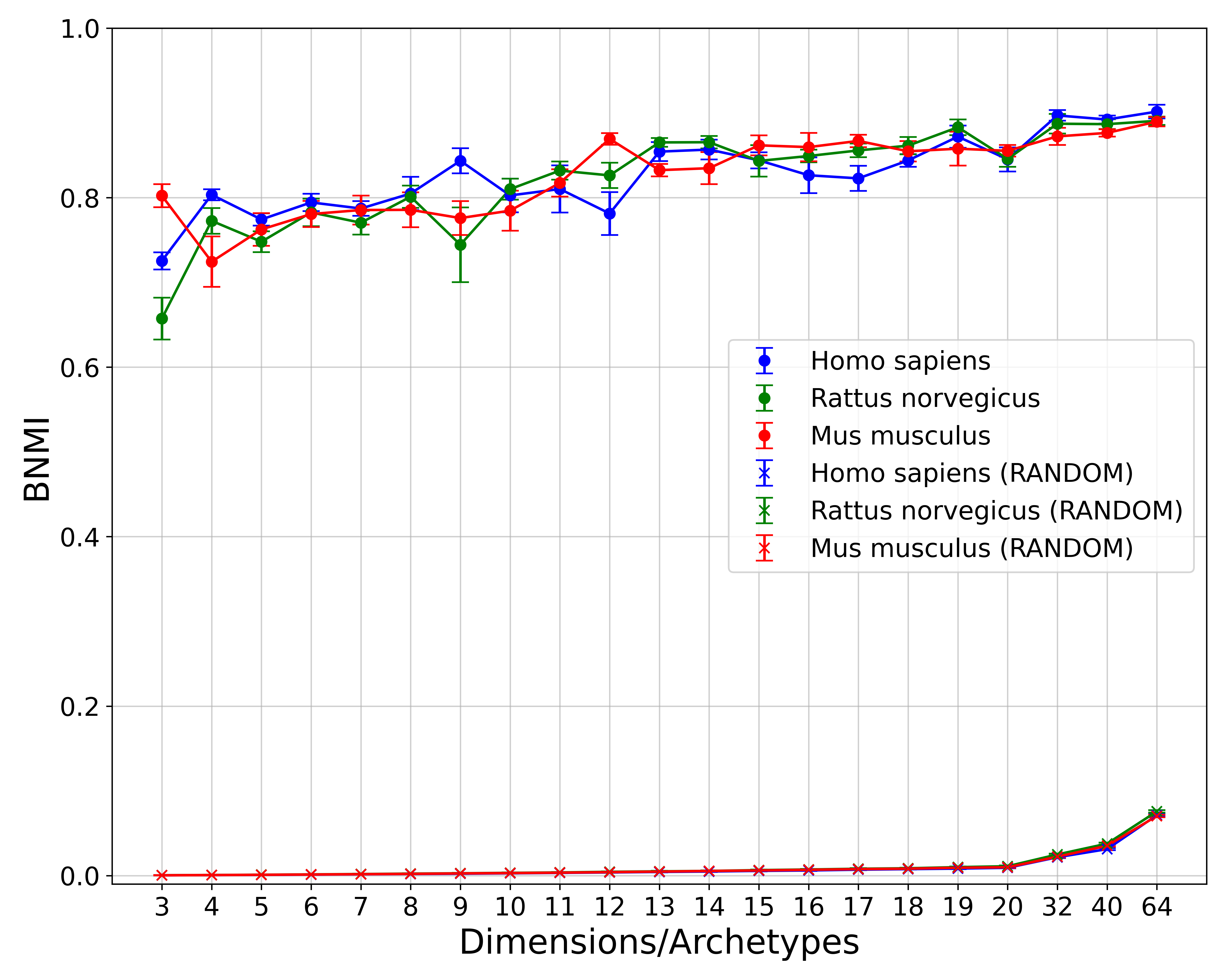}
    \caption{Positive Link Structure---$(\bm{Z} \text{ matrix})$}
    \label{fig:BNMI_Z}
\end{subfigure}
\hfill
\begin{subfigure}{0.49\columnwidth}
    \includegraphics[width=\textwidth]{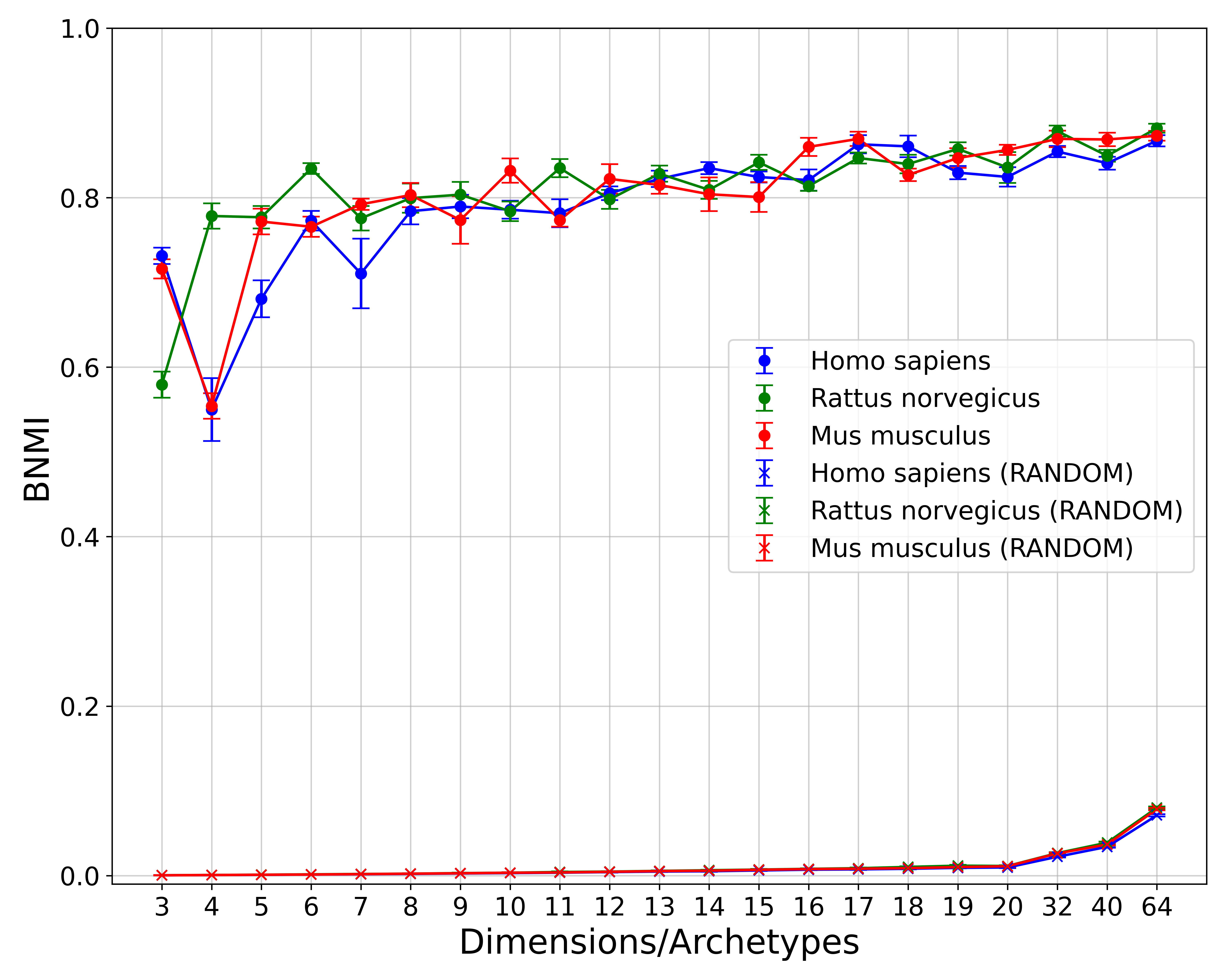}
    \caption{Negative Link Structure---$(\bm{W} \text{ matrix})$}
    \label{fig:BNMI_W}
\end{subfigure}
\caption{\textsc{Bayesian Normalized Mutual Information} (BNMI): Robustness of solution and structure characterization of \modelab, as a function of the number of dimensions/archetypes across five reruns, and three networks. For a given dataset, the (RANDOM) labeled lines denote the BNMI value of solution that should be expected by luck, for a given choice for the number of archetypes/dimensions.}
\label{fig:rob}
\end{figure}

Next, we evaluate the representation capabilities of the proposed model from quantitative and qualitative aspects. Specifically, we assess its effectiveness in link prediction and in successfully inferring positive and negative archetypes and communities, accompanied by enrichment analysis. 

\textbf{Robustness and Identifiability of the Solution:} An important aspect of a model characterizing the structure of a given signed network is the robustness, as well as the identifiability of the given solution. For that, we here present the results of the BNMI across five model reruns, as described in the previous section. Specifically, in Figure \ref{fig:BNMI_Z}, and Figure \ref{fig:BNMI_W}, we provide the BNMI scores for both the positive and negative space mixed-membership matrices $\bm{Z},\bm{W}$, respectively. Essentially, we compare the consistency of the soft assignments with respect to the archetypes across the different runs of the model. We also provide the BNMI scores that should be expected by luck, presented with (RANDOM) in the legends of the aforementioned Figures. We consider all three datasets for various dimensions, ranging from $3$ to $64$. For the positive space mixed-membership matrix $\bm{Z}$, we observe that the BNMI has a gradual and modest increase as the number of dimensions grows for all three datasets. For the (RANDOM) permutations, the BNMI scores are essentially zero for all dimensions of less than $32$, and show very small values for greater dimensions. For the negative space mixed-membership matrix $\bm{W}$, we notice the same behavior as in the positive case, with the main difference that in model configurations with fewer than $7$ archetypes, the BNMI score fluctuates more. Based on these results, we conclude that the model obtains consistent results and high \textsc{BNMI} scores across reruns when $K>4$ for the positive case and $K>6$ for the negative case. Thus, our \modelab yields successful and robust structure characterization. In the case of the randomly permuted solutions, the \textsc{BNMI} scores are $0$, validating that there is no structure retrieval due to randomness. To calculate the BNMI score that should be expected by luck, we have considered $100$ permutations of the solution for every, reporting the average scores across reruns.

\textbf{\modelab Significantly Outperforms Baselines in Signed Link Prediction:} To evaluate the predictive capability of our model, we consider the general task of signed link prediction. Contrary to previous works \citep{SLIM, slf, pole}, we extend the experimental setting beyond binarized sign prediction and link prediction. We instead formulate the task as a three-class classification problem (negative (\textit{neg}), zero (\textit{zr}), and positive (\textit{pos}) links). For this task, we remove (or set to zero) 10\% of the total network links while ensuring that the remaining network stays connected. We then train the model on this residual network and evaluate its ability to simultaneously predict both the sign and the existence of the removed links. Specifically, for a given pair of proteins, our goal is to determine whether they exhibit negative regulation (\textit{neg}), positive regulation (\textit{pos}), or no interaction at all (\textit{zr}). (Additional settings are provided in the supplementary.)

We compare our models' performance against seven prominent methods for signed networks. We present results regarding per class and total weighted ($w$) F1 scores in Table~\ref{tab:f1_score}. We here witness that \modelab outperforms, and in most cases significantly, all the considered baselines in all classes and on the weighted across classes F1 score. Table~\ref{tab:f1_score} further shows that the negative regulation class (\textit{neg}) is the most challenging to predict, highlighting the importance of models that account for this setting.  Regarding the baselines, \textsc{SLIM} is the most competitive method following \textsc{S$2$-SPM}, highlighting the superiority of latent distance modeling in link prediction. Importantly, only the latent distance models achieved high scores in all classes of the considered problem, with \textsc{SLF} having also favorable performance.. The rest of the baselines were observed to be competitive only to a subset of the tasks and thus can be characterized as non-robust in their predictive performance. Overall, the proposed \modelab demonstrated improved predictive performance for all three datasets.

\textbf{Enrichment Analysis of the Archetypes:}
To validate the biological relevance of the identified archetypes, we next analyze the GO terms that were found to be enriched in each archetype, focusing on the \textsl{Homo sapiens} network.

\textit{Down-regulation Archetypes Enrichment Analysis:} We found some identified archetypes to be enriched for proteins associated with distinct biological processes. Among these, archetype $5$ represents the processes of antiviral immune response: defence response to viruses, innate immune response, mRNA binding, positive regulation of interferon-alpha production, and positive regulation of type I interferon production. Specifically, it captures positive regulation of interferon-alpha, a type I interferon predominantly produced by innate immune cells in response to viral infection \citep{neg1,neg2}. Similarly, archetype $8$ captures key processes and components with common biological relevance integral to mitosis, including cell division, structural constituent of cytoskeleton, mitotic cell cycle, mitotic spindle, microtubule cytoskeleton, microtubule, microtubule cytoskeleton organisation, myosin phosphatase activity, and MAPK cascade \citep{neg3}. Collectively, this archetype captures the last phase of the cell cycle, a type of cell division essential for growth, development, and repair \citep{neg4,neg5}. Archetype $4$, is primarily involved in regulating Rho GTPase signalling, which is crucial in various biological processes, such as cell cytoskeletal organisation, differentiation, growth, neuronal development, and synaptic functions \citep{clus4_neg}. Other archetypes were more complex to interpret as they reflect multiple biological processes. Archetype $1$, is centered around cell migration, adhesion, and tissue organization, in the context of development, epithelial to mesenchymal transition (EMT) and extracellular matrix remodeling. Enrichment in the transforming growth factor (TGF)-$\beta$ signaling pathway, essential for EMT and migration, both known for their importance during development and cancer progression \citep{neg7,neg8}. Archetype $2$, is mainly associated with G protein-coupled receptor (GPCR) signaling, synaptic transmission, and cell signaling at the plasma membrane. Its enrichment in key roles in calcium signaling, neuropeptide signaling, and cell adhesion, indicates its involvement in modulating neuronal communication and cellular interactions. Archetype $3$, features its key role in cell-cell signaling, processes related to the extracellular space and matrix, and hormonal and ligand-receptor activities. Archetype $6$, is the smallest among the archetypes and seems to represent a dual function: one in mitochondrial protein synthesis and the other in chromatin structure. Archetype $7$, integrates processes related to autophagy, apoptosis, and cell division. It likely reflects the role of mitophagy, determining whether a cell under mitochondrial stress survives through repair (autophagy) or undergoes cell death (apoptosis) \citep{cl7}.

\textit{Up-regulation Archetypes Enrichment Analysis:} All positively regulated archetypes are relatively large, often encompassing diverse GO terms, which contribute to their complexity and the challenge of assigning a specific biological function, while also implying dynamic interactions among their components. Archetype $2$, stands out with a direct biological association, as it is predominantly linked to the cell cycle, with a strong emphasis on cytoskeletal organization, chromosome segregation, and DNA repair processes. Both archetypes $4$ and $7$ are primarily associated with protein homeostasis, with a strong emphasis on regulating protein degradation via ubiquitination \citep{pos1,pos2}. These seem to reflect the two major pathways for protein degradation: autophagy (archetype $4$) and ubiquitin–proteasome system (archetype $7$) \citep{auto5}. Archetype 5, is strongly associated with G protein-coupled protein activity and regulation, neurotransmission, hormone activity and calcium signaling, implying its role in synaptic signaling \citep{clus56}.  Archetype $8$, is the largest and encompasses both innate and adaptive immune responses and inflammation. The rest of the archetypes exhibit mixed biological signals with dynamic interplay including bone development and differentiation, TGF-$\beta$ signaling, morphogenesis, transcription and immune response regulation (archetype $1$); translation, cell cycle regulation, synaptic transmission and neural activity (archetype $3$); multiple processes among which chromatin organization and Wnt signaling (archetype $6$). 

In summary, the identified positive and negative space archetypes are often complex but reflect significant biological processes. This complexity is likely influenced by protein interaction signals captured across various cell types, each differentiated by the nature and levels of specific proteins they express. As UniProt integrates data from both healthy and diseased sources, it may already capture dysregulation in its proteome due to pathological conditions or has the potential to detect it (examples of the enrichment labels are provided in supplementary).

\textbf{PPI Network Visualizations:} We present visualizations obtained from the proposed \modelab, demonstrating its capability to extract informative and robust latent structures. Figure \ref{fig:polytopes_pos}, illustrates the inferred latent structures for both positive and negative spaces as defined by the model. Specifically, Figures \ref{fig:polytopes_pos}(a) and \ref{fig:polytopes_pos}(c), depict the projections of the positive and negative space latent embeddings, $\Tilde{\bm{Z}} = \bm{A}_{(+)}\bm{Z}$ and $\Tilde{\bm{W}} = \bm{A}_{(-)}\bm{W}$, onto circular plots. These are enriched with edges connecting nodes assigned to the same archetype. Each archetype is represented as a point evenly spaced around a circle, positioned every $\text{rad}_k = \frac{2\pi}{K}$ radians, where \(K\) is the total number of archetypes. The circular plots highlight \modelab’s ability to allocate nodes to distinct archetypes, revealing proteins behaving as archetypes or extreme profiles within the data. Figures \ref{fig:polytopes_pos}(b) and \ref{fig:polytopes_pos}(d) display the reordered adjacency matrices based on the archetype assignments of the positive and negative mixed-membership matrices, $\bm{Z}$ and $\bm{W}$. These matrices effectively uncover the underlying latent or block structures in the data. This visualization demonstrates \modelab's effectiveness in identifying and characterizing the archetypes and latent structures inherent in the data, offering insights into the underlying biological processes.

%% file: discussion.tex
\vspace{-3pt}
\section{Discussion}

Automatically predicting interactions in complex biological networks remains a challenging yet crucial step for solving several biological tasks, including decoding disease mechanisms and accurately determining therapeutic targets.
In this study, we proposed the  Signed Two-Space Proximity Model (\textsc{S2-SPM}), tailored to the machine learning modeling of signed PPI networks. Specifically, we introduced two latent spaces to decouple positive and negative network interactions, assuming that both interaction types should be translated into close proximity in a latent space model. Prominent modeling techniques for PPI networks are typically blind to the sign of interactions, a limitation addressed by the proposed model. 

Additionally, our method addresses the circularity concerns that commonly arise in classical archetypal analysis studies \citep{AA_nmeth}, where the same data are used both to define the archetypes and to identify the enriched labels or traits within each archetype. In contrast, \textsc{S2-SPM} leverages the signed PPI network to identify the archetypal structure. This process is independent of each protein's underlying labels or GO terms. These labels are only used later for enrichment analysis. As a result, the two stages are decoupled, eliminating the risk of data leakage and the need for additional validation steps \citep{AA_nmeth}. This design choice further highlights the superiority of our method.

In experiments, \textsc{S2-SPM} outperformed all baselines, particularly in F1 scores for the signed link prediction task. Notably, our model significantly surpassed recent signed network models based on latent distance and the Skellam distribution \citep{SLIM}, emphasizing the importance of using two independent latent spaces for modeling each interaction type in signed protein-protein networks.
Furthermore, we presented that the obtained archetype structures could further be enriched with the GO terms characterizing the different proteins present in the \textsl{homo-sapiens} network. 
Specifically, we showcased that both positive and negative interactions formed archetypal groups carrying out different biological tasks.
The obtained archetype structures were also tested for statistical significance, robustness, and spurious structure retrieval. Comparisons with information signals given by chance under random permutations of the archetypal membership matrix also confirmed that our model yields reliable and consistent structures for both latent spaces.
This analysis proves that \textsc{S2-SPM} constitutes an identifiable approach for modeling complex protein interactions while ensuring that key biological features are precisely captured and interpretable. 